\documentclass{article}

    \PassOptionsToPackage{numbers, compress}{natbib}

\usepackage[final]{neurips_2024}

\usepackage[dvipsnames]{xcolor}

\usepackage{algpseudocode, algorithm}
\usepackage{amsmath,amsthm,amssymb,mathtools}
\usepackage{subcaption}
\usepackage{comment}
\usepackage{multirow}
\usepackage{tikz}
\usepackage{bchart}
\usepackage{pgfplots}
\usepackage{bbding}
\usepackage{enumitem}

\newtheoremstyle{funny}
  {}{}
  {\itshape}
  {}
  {\bfseries}
  {.}
  { }
  {%
   \thmname{#1}
   \thmnumber{ #2}
   \thmnote{ {\mdseries\iffunny(\fi#3\iffunny)\fi}}
   \global\funnytrue 
  }
\newif\iffunny

\theoremstyle{funny}
\newtheorem{remark}{Remark}

\theoremstyle{funny}
\newtheorem{theorem}{Theorem}

\newtheorem{lemma}{Lemma}

\algnewcommand\algorithmicswitch{\textbf{switch}}
\algnewcommand\algorithmiccase{\textbf{case}}
\algnewcommand\algorithmicassert{\texttt{assert}}
\algnewcommand\Assert[1]{\State \algorithmicassert(#1)}%
\algdef{SE}[SWITCH]{Switch}{EndSwitch}[1]{\algorithmicswitch\ #1\ \algorithmicdo}{\algorithmicend\ \algorithmicswitch}%
\algdef{SE}[CASE]{Case}{EndCase}[1]{\algorithmiccase\ #1}{\algorithmicend\ \algorithmiccase}%
\algtext*{EndSwitch}%
\algtext*{EndCase}%

\newcommand{\cmark}{\textcolor{green}{\Checkmark}}
\newcommand{\xmark}{\textcolor{red}{\XSolidBrush}}

\usepackage{colortbl}
\definecolor{bGreen}{HTML}{D3FFD3}
\definecolor{bBlue}{HTML}{D3D3FF}
\definecolor{bRed}{HTML}{FFD3D3}
\definecolor{bGray}{HTML}{D3D3D3}

\definecolor{boxred}{rgb}{1,0.7,0.7}
\definecolor{boxgreen}{rgb}{0.7,1,0.7}

\definecolor{half_magenta}{rgb}{1, 0.7, 1}
\definecolor{half_cyan}{rgb}{0.7, 1, 1}

\definecolor{tabfirst}{rgb}{1, 0.7, 0.7} 
\definecolor{tabsecond}{rgb}{1, 0.85, 0.7} 
\definecolor{tabthird}{rgb}{1, 1, 0.7} 

\newcommand{\ie}{\textit{i.e.}}
\newcommand{\eg}{\textit{e.g.}}

\usepackage{helvet}
\usepackage[eulergreek]{sansmath}
\pgfplotsset{
    compat=1.17,
    every axis/.append style={
        title={\small\sffamily #1}, 
        xlabel={\small\sffamily #1}, 
        ylabel={\small\sffamily #1}, 
        legend style={font=\small\sffamily}, 
        tick label style={font=\small\sffamily}, 
        title style={font=\small\sffamily}, 
        xticklabel style={font=\sansmath\sffamily}, 
        yticklabel style={font=\sansmath\sffamily} 
    },
    every axis label/.append style={font=\small\sffamily}
}

\newcommand{\plotblack}[1]{%
    \addplot[
        color=black,
        error bars/.cd,
        y dir=both,
        y explicit,
        error bar style={line width=1pt},
        error mark options={
            rotate=90,
            mark size=5pt,
            line width=0.1pt
        },
    ]
    coordinates {#1};

    \addplot[color=black ,mark=square, only marks, mark options={
        fill=none,
        draw=black
    }] coordinates {#1};
}

\newcommand{\plotmagenta}[1]{%
    \addplot[
        color=magenta,
        error bars/.cd,
        y dir=both,
        y explicit,
        error bar style={line width=1pt},
        error mark options={
            rotate=90,
            mark size=5pt,
            line width=0.1pt
        },
    ]
    coordinates {#1};

    \addplot[color=magenta,mark=triangle, only marks] coordinates {#1};
}

\newcommand{\plotcyan}[1]{%
    \addplot[
        color=cyan,
        error bars/.cd,
        y dir=both,
        y explicit,
        error bar style={line width=1pt},
        error mark options={
            rotate=90,
            mark size=5pt,
            line width=0.1pt
        },
    ]
    coordinates {#1};

    \addplot[color=cyan,mark=*, only marks] coordinates {#1};
}

\usepackage{amsmath,amsfonts,bm}

\def\eqref#1{equation~\ref{#1}}

\def\1{\bm{1}}

\def\eps{{\epsilon}}

\def\vv{{\bm{v}}}
\def\vw{{\bm{w}}}
\def\vx{{\bm{x}}}
\def\vy{{\bm{y}}}
\def\vz{{\bm{z}}}

\DeclareMathAlphabet{\mathsfit}{\encodingdefault}{\sfdefault}{m}{sl}
\SetMathAlphabet{\mathsfit}{bold}{\encodingdefault}{\sfdefault}{bx}{n}

\def\gD{{\mathcal{D}}}
\def\gE{{\mathcal{E}}}

\def\gG{{\mathcal{G}}}

\def\gI{{\mathcal{I}}}

\def\gN{{\mathcal{N}}}

\def\gT{{\mathcal{T}}}

\newcommand{\R}{\mathbb{R}}

\usepackage[utf8]{inputenc} 
\usepackage[T1]{fontenc}    
\usepackage{hyperref}       

\usepackage[capitalize]{cleveref}
\usepackage{url}            
\usepackage{booktabs}       
\usepackage{amsfonts}       
\usepackage{nicefrac}       
\usepackage{microtype}      
\usepackage{xcolor}         
\newcommand*\samethanks[1][\value{footnote}]{\footnotemark[#1]}

\title{Gradient-free Decoder Inversion \\ in Latent Diffusion Models}

\author{%
  \vspace{5pt} 
  Seongmin Hong$^{1}$ \quad \quad \quad \quad Suh Yoon Jeon$^{1}$ \quad \quad \quad \quad Kyeonghyun Lee$^{1}$ \\ \vspace{5pt}
  \textbf{Ernest K. Ryu}$^{2,}$\thanks{Co-corresponding authors} \quad \quad \quad \quad \textbf{Se Young Chun}$^{1,3,}$\samethanks[1]\\ 
  $^1$Dept. of Electrical and Computer Engineering, $^3$INMC \& IPAI, Seoul National University\\ $^2$Dept. of Mathematics, University of California, Los Angeles\\
  \texttt{\{smhongok, euniejeon, litiphysics, sychun\}@snu.ac.kr,} \texttt{eryu@math.ucla.edu}\\
}

\begin{document}

\maketitle

\begin{abstract}
In latent diffusion models (LDMs), denoising diffusion process efficiently takes place on latent space whose dimension is lower than that of pixel space. Decoder is typically used to transform the representation in latent space to that in pixel space. While a decoder is assumed to have an encoder as an accurate inverse, exact encoder-decoder pair rarely exists in practice even though applications often require precise inversion of decoder. Prior works for decoder inversion in LDMs employed gradient descent inspired by inversions of generative adversarial networks. However, gradient-based methods require larger GPU memory and longer computation time for larger latent space. For example, recent video LDMs can generate more than 16 frames, but GPUs with 24 GB memory can only perform gradient-based decoder inversion for 4 frames. Here, we propose an efficient gradient-free decoder inversion for LDMs, which can be applied to diverse latent models. Theoretical convergence property of our proposed inversion has been investigated not only for the forward step method, but also for the inertial Krasnoselskii-Mann (KM) iterations under mild assumption on cocoercivity that is satisfied by recent LDMs. Our proposed gradient-free method with Adam optimizer and learning rate scheduling significantly reduced computation time and memory usage over prior gradient-based methods and enabled efficient computation in applications such as noise-space watermarking while achieving comparable error levels.
\end{abstract}

\section{Introduction}
Deep generative models have been actively investigated over the past decade across numerous modalities such as image~\cite{ho2020denoising,song2020denoising,song2020improved,rombach2022high,ho2021classifierfree,ramesh2022hierarchical,ruiz2023dreambooth}, video~\cite{blattmann2023align,blattmann2023stable,wang2023lavie,zhou2022magicvideo,wang2023modelscope,wang2023videolcm}, audio~\cite{kong2020diffwave,liu2023audioldm,liu2023audioldm2} and molecular structure~\cite{kohler2021smooth, hong2023neural}. They can sample new data points from the distribution of the training set, can model priors to solve regression problems, and can be used for applications like editing, retrieval, and density estimation.
Representative classes of deep generative models include generative adversarial networks (GANs), variational autoencoders (VAEs), normalizing flows (NFs), and diffusion models (DMs). They \emph{used to be} known not to simultaneously satisfy the three key requirements: (i) high-quality samples, (ii) diversity, and (iii) fast sampling, also referred to as the generative learning trilemma~\citep{xiao2021tackling}. However, recent DMs such as Rectified Flow~\citep{liu2022flow}, Adversarial Diffusion Distillation~\citep{sauer2023adversarial} and Consistency models~\citep{song2023consistency} require just one step for sampling. Thus, DMs have overcome the trilemma and the old problem of slow sampling in recent DMs is no longer a concern as shown in the last row of \Cref{tab:generative_models}. As a result, DMs have become one of the most prominent deep generative models, especially for image and video. However, there are still remaining challenges for DMs such as achieving \emph{invertibility}, which is well-supported by other models (see \Cref{tab:generative_models}).


\begin{table}[!t]
\caption{Comparison of deep generative models. DMs have overcome the \emph{generative learning trilemma}~\citep{xiao2021tackling}, but they still lack invertibility compared to other models. In particular, LDMs have necessitated additional decoder inversion that has traditionally been addressed through memory-intensive and time-consuming gradient-based optimization methods. Here, we propose a method for efficiently (\ie, gradient-free) ensuring invertibility in LDM.}
\label{tab:generative_models}
\small
\centering
\begin{tabular}{lcccc}
\toprule
                                                    & \multicolumn{3}{c}{Generative learning trillema~\citep{xiao2021tackling}}                                                                                                                                                                     &                    
\\
\multicolumn{1}{l|}{}                               & \begin{tabular}[c]{@{}c@{}}High-quality\\ samples\end{tabular} & \begin{tabular}[c]{@{}c@{}}Diversity\end{tabular} & \multicolumn{1}{c|}{\begin{tabular}[c]{@{}c@{}}Fast \\ sampling\end{tabular}} & 
\begin{tabular}[c]{@{}c@{}}How to achieve \\ invertibility?\end{tabular} \\ \midrule
\multicolumn{1}{l|}{GAN}                            & \cmark                                          & \xmark                                               & \cmark                                                         & Optimization / learning-based GAN inversion \\ \cmidrule{5-5}
\multicolumn{1}{l|}{NF}               & \xmark                                          & \cmark                                               & \cmark                                                         & Naturally exact inversion\\ \cmidrule{5-5}
\multicolumn{1}{l|}{DM} & \cmark                                          & \cmark                                               & \xmark$_{\text{('21)}}$ $\rightarrow$ \cmark$_{\text{('24)}}$                                                        & \begin{tabular}[c]{@{}c@{}} Diffusion: Solving ODE backward \\ \cellcolor[HTML]{FFFC9E} (LDM) Decoder: Optimization-based inversion  \end{tabular}              \\ \bottomrule 
\end{tabular}
\end{table}

Achieving invertibility for deep generative models has been an important topic. In GANs, prior works proposed GAN inversion techniques~\citep{xia2022gan} (\ie, to find the latent vector $\vz$ that generated the image) and their applications such as real image editing~\citep{abdal2019image2stylegan}. NFs are naturally invertible deep generative models by construction with related interesting applications~\citep{kohler2021smooth}. There have also been attempts to achieve and utilize invertibility for DMs. The most popular one is the na\"ive DDIM inversion that reverses the sampling process of DDIM's deterministic denoising~\cite{song2020denoising}, \ie, adding the estimated noise. While the na\"ive DDIM inversion enables image editing, it is known to be somewhat inaccurate for seeking the true latent~\citep{hong2023exact}, for watermarking~\citep{wen2023tree} and for background-preserving image editing~\citep{patashnik2023localizing}. To better ensure the exactness of the inversion, several prior works~\cite{wallace2023edict, zhang2023exact, Pan_2023_ICCV, hong2023exact, garibi2024renoise} have proposed more tailored algorithms than the na\"ive DDIM inversion.

The use of latents in DMs has made ensuring invertibility more difficult. Latent diffusion models (LDMs) were proposed to move the diffusion denoising process from the pixel space to the (low-dimensional) latent space, thus efficiently generating high-quality and large-scale samples~\citep{rombach2022high}. This issue may be critical since many recent popular DMs operate in latent space~\citep{rombach2022high, wang2023lavie, zhou2022magicvideo, luo2023latent, sauer2023adversarial, wang2023modelscope, blattmann2023align}. However, latents are usually lossy compression of pixels, making one-to-one mapping between the latent and pixel spaces very challenging.
Thus, the accuracy of inversion in LDMs is typically lower than that in pixel-space~\citep{wallace2023edict, Pan_2023_ICCV, hong2023exact}, requiring additional efforts to compensate for it as illustrated in~\Cref{fig:generative_models}.
One could na\"ively employ a gradient-based GAN inversion method to the decoder of LDMs, but it required very large GPU memory and computation time~\citep{hong2023exact}, especially for large-scale LDMs.
Moreover, recent video LDMs~\cite{blattmann2023align,blattmann2023stable,wang2023lavie,zhou2022magicvideo,wang2023modelscope,wang2023videolcm} generate dozens of frames at once, making gradient-based GAN inversion infeasible in a single GPU.

In this work, we aim to achieve better invertibility in LDM as shown in~\Cref{tab:generative_models}. Specifically, we propose decoder inversion to overcome the difficulty of ensuring invertibility in LDM due to the inexact encoder-decoder pair. The proposed decoder inversion is gradient-free, which is faster and more memory-efficient than the gradient-based methods suggested in GAN inversion. \Cref{sec:method} describes the motivation and analysis of our decoder inversion, including the theoretical convergence guarantee under mild assumption not only for the vanilla forward step method, but also for the inertial Krasnoselskii-Mann (KM) iterations. The assumption for our theorems is also validated by showing experimental convergence to the ground truth as well as the effectiveness of momentum. \Cref{sec:experiments} described how to refine our algorithm by integrating popular optimization techniques such as Adam~\cite{kingma2014adam} and learning rate scheduling as practical extensions of our gradient-free decoder inversion, demonstrating that our method works well on various latest LDMs compared to existing gradient-based methods. Lastly, \Cref{sec:app} showcases interesting application where our proposed gradient-free decoder inversion can be effectively utilized. The contributions of this work are:
\begin{itemize}
    \item proposing a gradient-free decoder inversion algorithm to achieve better invertibility in diverse latest LDMs. Our method has the following advantages with experimental evidences: 
        \begin{itemize}
            \item \textbf{Fast}: up to 5$\times$ faster, 1.89 s vs 9.51 s to achieve -16.4 dB (Fig. 3c and Tab. S1c)
            \item \textbf{Accurate}: up to 2.3 dB lower, -21.37 dB vs -19.06 dB in 25.1 s (Fig. 3b and Tab. S1b)
            \item \textbf{Memory-efficient}: up to \textbf{89\%} can be saved, 7.13 GB vs 64.7 GB (Fig. 3b)
            \item \textbf{Precision-flexible}: 16-bit vs 32-bit (Fig. 3)
        \end{itemize}
    \item theoretically guaranteeing the convergence to the ground truth not only for the vanilla forward step method, but also for the inertial KM iterations,
    \item showcasing that our proposed gradient-free decoder inversion can be used in interesting applications.
\end{itemize}

\section{Backgrounds}\label{sec:backgrounds}
\subsection{Latent diffusion models (LDMs)}\label{sec:ldms}

Stable Diffusion 2.1~\citep{rombach2022high} is a widely known open-source text-to-image generation model. LaVie~\citep{wang2023lavie} is a text-conditioned video generation model, which can generate consecutive frames per inference. InstaFlow~\citep{liu2023instaflow} is a one-step text-to-image generation model that can generate images whose quality is as good as Stable Diffusion. These three LDMs will be used in all the experiments of our work.

\paragraph{Exact inversion of accelerated DMs.}
Another factor that makes the exact inversion of DMs difficult, besides the use of latents, is acceleration. Due to the use of high-order ODE solvers, the exact inversion of DPM-Solvers~\citep{lu2022dpm,lu2022dpm++} is not as straightforward as in DDIM. Exact inversion algorithms for some DPM-Solvers have been proposed~\citep{hong2023exact}, but obtaining the exact inverses for other accelerated DMs~\citep{liu2022flow, sauer2023adversarial, song2023consistency, liu2023instaflow} has not been explored. 

\begin{figure}[!t]
    \centering
    \begin{subfigure}{0.35\textwidth}
        \centering
        \begin{tikzpicture}
            \node[inner sep=0pt] at (0,0) {\includegraphics[width=\textwidth]{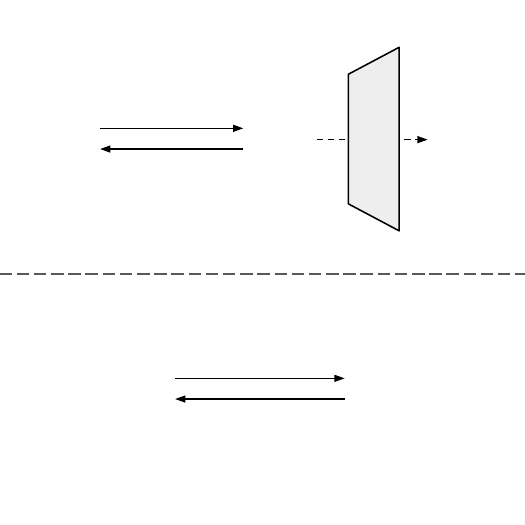}};
            \node[font=\fontsize{10}{9}\selectfont] at (-0.8, 1.5) {Diffusion};
            \node at (-1.9,1.15) {$\vz_T$};
            \node[font=\fontsize{9}{9}\selectfont] at (-1.9,0.8) {(Noise)};
            \node at (0.2,1.15) {$\vz_0$};
            \node at (1.05,1.15) {$\gD$};
            \node at (1.8,1.15) {$\vx$};
            \node[font=\fontsize{9}{9}\selectfont] at (1.85,0.8) {(Image)};

            \node at (0,0.1) {LDM};
            \node at (0,-0.4) {Pixel-space DM};
            \node at (-1.2,-1.2) {$\vz_T$};
            \node[font=\fontsize{9}{9}\selectfont] at (-1.2,-1.55) {(Noise)};
            \node at (1.1,-1.2) {$\vz_0$};
            \node[font=\fontsize{9}{9}\selectfont] at (1.1,-1.55) {(Image)};
        \end{tikzpicture}
        \caption{LDM vs Pixel-space DM}
    \end{subfigure}
    \begin{subfigure}{0.63\textwidth}
        \centering
        \large
        \begin{tikzpicture}
            \node[inner sep=0pt] at (0,0) {\includegraphics[width=\textwidth]{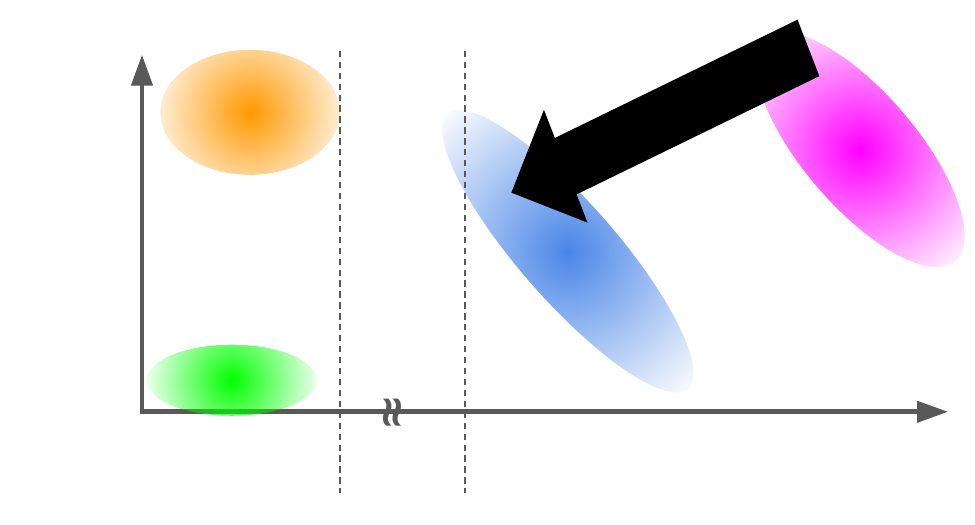}};
            \node[rotate=26,text=yellow,font=\fontsize{10}{15}\selectfont\sffamily] at (1.6,1.3) {Our contribution};

            \node at (-2.3,-1.05) {NF};
            \node at (-2.2, 1.7) {VAE};

            \node[align=center, font=\fontsize{9}{9}\selectfont] at (-2, 1.2) {Learning-based \\ GAN inversion};
            
            \node[align=left, font=\fontsize{9}{9}\selectfont] at (1.6, -0.3) {Pixel-space DM\\Opt.-based GAN inversion};

            \node at (3.4, 1) {LDM};

            \node[align=left,font=\fontsize{10}{10}\selectfont] at (-3.3, 1.9) {Inversion\\Error};

            \node[align=left,font=\fontsize{10}{10}\selectfont] at (3.4, -1.8) {Inversion\\Runtime};

            \node[font=\fontsize{9}{9}\selectfont] at (-2.5, -1.8) {1-step grad-free};

            \node[font=\fontsize{9}{9}\selectfont] at (1.1, -1.8) {Iterative grad-based};

        \end{tikzpicture}
        \caption{Invertibility of generative models}
        \label{fig:generative_models}
    \end{subfigure}
    \caption{(a) The difference between LDM and pixel-space DM lies in the use of the decoder ($\gD$). (b) This difference has caused the performance gap in exact inversions. 
    Pixel-space DMs and gradient-based GAN inversion methods are located on the right side due to iterative gradient back-propagations. If gradient-based decoder inversions are used, which are computationally intensive, LDM is located on the rightmost side to address the lossiness of latents. Our proposed gradient-free decoder inversion method allows us to efficiently handle the transformation between latent and pixel spaces.
    }\label{fig:1}
\end{figure}
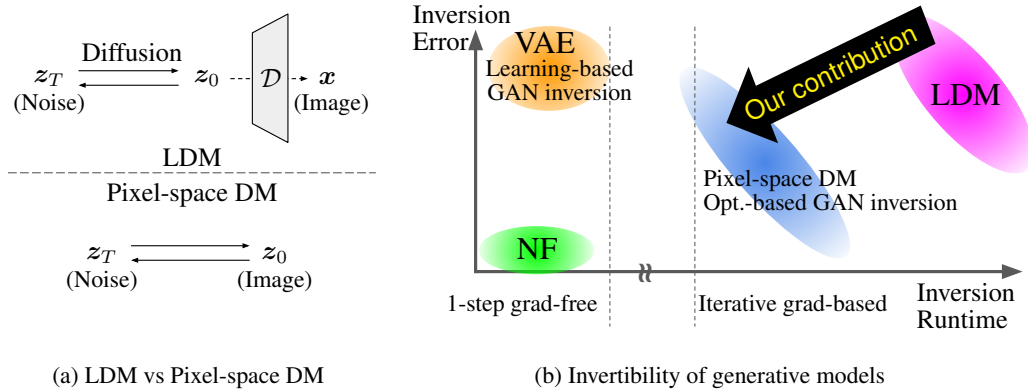

\subsection{Optimization-based GAN inversion}\label{sec:gan_inversion}
For a target image $\vx$, existing optimization-based GAN inversion methods typically perform the following optimization with respect to a latent vector $\vz$ (usually assumed to lie on a simple distribution such as $\gN(0,1)$):
\begin{equation}
\underset{\vz}{\mathrm{min}} \quad \ell(\vx, \gG(\vz)),
\label{eqn:opt}
\end{equation}
where $\ell$ is a distance metric and $\gG$ is the generator part of GAN~\cite{xia2022gan}. Most works solve \cref{eqn:opt} through backpropagation with an optimizer. For optimizers, Adam~\citep{kingma2014adam} is employed in \citep{abdal2019image2stylegan}, and L-BFGS \citep{liu1989limited} is employed in \citep{zhu2016generative}. Some more recent works optimized the transformed latent feature $\vw$, which can be defined as $\vw = \text{MLP}(\vz)$, in StyleGAN~\citep{karras2019style}.

To exploit the prior knowledge that $\vz$ lies on the range space of an encoder, prior works often perform the following two methods:
\begin{enumerate}[label=(\Alph*)]
    \item The output of the encoder $\gE$ is used as an initial point of the optimization process. \label{2.2.a}
    \item The domain-guided encoder is used to regularize the latent code within the semantic domain of the generator. For example, $\underset{\vz}{\mathrm{min}} \; \|x-\gG(\vz)\|_{2} +\lambda_{\text{reg}}\|\vz-\gE(\gG(\vz))\|_{2}$. \label{2.2.b}
\end{enumerate}
Since we already have the encoder, we employ \ref{2.2.a}, too. On the other hand, we will utilize the encoder in a novel way over \ref{2.2.b}.

\subsection{Gradient-based decoder inversion in LDMs}
Recent works~\cite{hong2023exact} performed the same as \cref{eqn:opt} (hence they are iterative as illustrated in \Cref{fig:generative_models}) for the decoder inversion:
\begin{equation}
\underset{\vz_0}{\mathrm{min}} \quad  \ell(\vx, \gD(\vz_0)),
\label{eqn:opt_z0}
\end{equation}
with a gradient-based method. For simplicity, we omit the subscript by setting $\vz_0 = \vz$. The vanilla gradient descent algorithm can be expressed as follows:
\begin{equation}
    \vz^{k+1} =\vz^{k} - \rho \nabla_{\vz} \ell (\vx, \gD(\vz^k)).
\label{eqn:gd}
\end{equation}
Advancements in backpropagation algorithms and libraries have made the computation of $\nabla_{\vz} \gD(\vz^k)$ more efficient, but it still requires a significant amount of GPU memory usage and lengthy runtime compared to inference. As the size of generated outputs of recent LDMs continues to increase, the computation of $\nabla_{\vz} \gD(\vz^k)$ through backpropagation is getting more and more burdensome.

\section{Gradient-free decoder inversion in LDMs}\label{sec:method}

In this work, as an alternative method to \cref{eqn:gd}, we propose the following decoder inversion method, which is a form of the 
forward step method~\citep{ryu2022large}: 
\begin{equation}
    \vz^{k+1} =\vz^{k} - \rho (\gE(\gD(\vz^k)) - \gE(\vx)),
\label{eqn:fsm}
\end{equation}
where the initial latent vector estimate $\vz^0 = \gE(\vx)$ that is the same as \ref{2.2.a} in \Cref{sec:gan_inversion}.

\subsection{Motivation}\label{sec:motivation} The definition of the decoder inversion problem for the given image $\vx$ is as follows:
\begin{equation}
    \underset{\vz \in \R^F}{\mathrm{find}} \quad \vx = \gD(\vz).
\label{eqn:find_pixel}
\end{equation}
Since directly dealing with \cref{eqn:find_pixel} is impractical, we relax it as solving the following problem:
\begin{equation}
    \underset{\vz \in \R^F}{\mathrm{find}} \quad \gE(\vx) = \gE(\gD(\vz)).
\label{eqn:find_latent}
\end{equation}
The following Remark 1 demonstrates that solving \cref{eqn:find_latent} is easier than solving \cref{eqn:find_pixel}.
\begin{remark}[(\cref{eqn:find_latent} is easier than \cref{eqn:find_pixel})]\label{rem1}
$\{ \vz | \vx = \gD(\vz) \} \subset \{ \vz | \gE(\vx) = \gE(\gD(\vz)) \}$.    
\end{remark}
\Cref{rem1} is true because $\vx = \gD(\vz) \Rightarrow \gE(\vx) = \gE(\gD(\vz))$.    
\Cref{eqn:find_latent} is equivalent to the following:
\begin{equation}
    \underset{\vz \in \R^F}{\mathrm{find}} \quad \vz = \vz - \rho ( \gE(\gD(\vz)) - \gE(\vx) ), \quad \forall \rho \in \R \cap \{0\}^C
\label{eqn:find_latent_fpi}
\end{equation}
where \cref{eqn:find_latent_fpi} refers to finding a fixed point of an operation $\gI - \rho(\gE(\gD(\cdot)) - \gE(\vx))$ whose another form is \cref{eqn:fsm}, the forward step method.

\subsection{Convergence analysis on forward step method}\label{sec:analysis}
Here, we demonstrate that our method converges to a fixed-point under reasonable conditions. Furthermore, although we solved an easier problem represented by \cref{eqn:find_latent} rather than directly solving \cref{eqn:find_pixel}, we show that our proposed method surprisingly finds the true solution of \cref{eqn:find_latent}.

\begin{theorem}[Convergence of the forward step method]\label{thm:1} Let $\beta>0$, $0<\rho<2\beta$, and $\vx \in \R^N$. Assume $\gT(\cdot)=\gE \circ \gD(\cdot) - \gE(\vx)$ is continuous. Consider the iteration
\begin{equation}
    \vz^{k+1} = \vz^k - \rho \gT \vz^k \quad \quad \text{for} \quad k=0,1,\dots
\end{equation}
Assume $\vz^\star$ is a zero of $\gT$ (\ie, $\gT \vz^\star = 0$) and
\begin{equation}\label{eqn:cocoercivity}
    \langle \gT \vz^k, \vz^k - \vz^\star \rangle \geq \beta \lVert \gT \vz^k \rVert_2^2 \quad \text{for} \quad k=0,1,\dots
\end{equation}
Then, $\gT \vz^k \rightarrow 0$. If, furthermore, $\vz^k \rightarrow \vz^\infty$, then $\vz^\infty$ is a zero of $\gT$ (\ie, $\gT \vz^\infty = 0$).
\end{theorem}

\begin{proof}[Proof outline]
\Cref{eqn:cocoercivity} makes $\lVert \vz^{k+1} - \vz^\star \rVert_2^2 \leq \lVert \vz^{k} - \vz^\star \rVert_2^2 - \rho (2\beta - \rho) \lVert \gT \vz^k \rVert_2^2$. Then, sum for $k=0,\dots,\infty$.
\end{proof}

\Cref{thm:1} assumes that \cref{eqn:cocoercivity} holds, which is refered as $\beta$-cocoercivity.
The assumption makes sense because we expect $\gE \gD \simeq \gI$. 
In fact, for linear autoencoders, it is proven by \citet{baldi1989neural} that $\gE \gD = \gI$ as the following \cref{rem2}:
\begin{remark}[$ED=\gI$ in linear autoencoders]\label{rem2}
Let $\vx \in \mathbb{R}^N$ be a random vector such that $\mathbb{E}[\vx \vx^\intercal]=\Sigma\in \mathbb{R}^{N\times N}$. Assume $\Sigma$ has distinct positive eigenvalues.
Consider the optimization problem
\[
\begin{array}{ll}
\underset{D\in \mathbb{R}^{N\times F},\,E\in \mathbb{R}^{F\times N}}{\mbox{min}}
&\mathop{\mathbb{E}}_{\vx}\|\vx-DE\vx\|^2_2.
\end{array}
\]
Then, $ED=I$ (identity matrix).
\end{remark}

$\gI$ is $1$-cocoercive, which suggests that our assumption is reasonable. In \cref{sec:assumption}, we further demonstrate that the assumption is reasonable experimentally and $\vz^k$ actually converges well. 

\subsection{Convergence analysis on momentum for acceleration}
Momentum is widely used in optimization to keep the optimization process going in the right direction, even when gradients are noisy or the landscape is flat. 
Among many momentum algorithms, we employ the inertial Krasnoselskii-Mann (KM) iteration~\cite{krasnosel1955two,mann1953mean,maulen2024inertial} and analyze it since the convergence is guaranteed with some assumptions~\cite{maulen2024inertial}. 
The inertial KM iteration in our setting can be defined as follows:
\begin{align}
    \vy^{k} & = \vz^k + \alpha(\vz^k - \vz^{k-1}) \label{eqn:momentum_1}\\
    \vz^{k+1} & = \vy^k -2 \lambda \beta \gT \vy^{k}, \label{eqn:momentum_2}
\end{align}
where $\gT(\cdot)=\gE \circ \gD(\cdot) - \gE(\vx)$, $0<\alpha<1$, $\beta>0$, and $\lambda > 0$.

We now present \Cref{thm:2}, ensuring that the inertial KM iterations converge. While the formulation and assumptions differ from \cite{maulen2024inertial}, the theorem and its proof heavily rely on it.
\begin{theorem}[Convergence of the inertial KM iterations]\label{thm:2} Let $0<\alpha<1$, $\beta>0$, $\lambda > 0$ and $\vx \in \R^N$. Assume $\gT(\cdot)=\gE \circ \gD(\cdot) - \gE(\vx)$ is continuous. Let $(\vz^k, \vy^k)$ satisfy (\ref{eqn:momentum_1}) and (\ref{eqn:momentum_2}). 
Assume $\vz^\star$ is a zero of $\gT$ (\ie, $\gT \vz^\star = 0$) and the following holds:
\begin{equation}\label{eqn:cocoercivity_y}
    \langle \gT \vy^k, \vy^k - \vz^\star \rangle \geq \beta \lVert \gT \vy^k \rVert_2^2 \quad \text{for} \quad k=0,1,\dots
\end{equation}
If
\begin{equation}\label{eqn:T2:0}
    \lambda (1 - \alpha + 2 \alpha^2) < (1-\alpha)^2,
\end{equation}
then the followings are true:
\begin{enumerate}[label=(\Alph*)]
    \item $ \sum_{k \geq 1} \|\vz^{k+1} -2\vz^k + \vz^{k-1} \|^2 $, $ \sum_{k \geq 1} \|\vz^k - \vz^{k-1} \|^2 $
    and $\sum_{k \geq 1} \|\gT \vy^{k} \|^2 $ converge. \label{thm2:conclusion:1}
    \item There is a constant $M>0$ such that $\min_{1\leq k \leq n} \|\gT \vy^{k} \|^2 \leq \frac{M \|\vz^1 - \vz^\star\|^2}{n}$. \label{thm2:conclusion:2}
    \item $\lim_{k\rightarrow\infty}\|\vz^k - \vz^\star\|$ exists. \label{thm2:conclusion:3}
    \item $(\vy^k)$ and $(\vz^k)$ have the same limit points. \label{thm2:conclusion:4}   
\end{enumerate}
\end{theorem}
\begin{proof}
See the supplementary material for the full proof.
\end{proof} 

Here, similar to \Cref{thm:1}, the convergence of $(\vz^k)$ is guaranteed for a learning rate (\ie, $2\lambda\beta$) smaller than some upper bound. However, unlike \Cref{thm:1}, this upper bound is determined not only by cocoercivity (\ie, $\beta$), but also by the inertial parameter (\ie, $\alpha$).

\subsection{Validation of the assumption}\label{sec:assumption}
In \Cref{thm:1,thm:2}, we utilized the assumption that $\gT(\cdot) = \gE \circ \gD (\cdot) - \gE(\vx)$ is $\beta$-cocoercive for the $(\vy^k, \vz^k)$ and $\vz^\star$. In this section, we experimentally investigate whether this assumption holds for the actual $(\vz^k)$ and $\vz^\infty$ obtained in the experiments of \Cref{sec:experiments}, \ie, 
\begin{equation}\label{eqn:actual_checking}
    \frac{\langle \gE \gD \vz^{\infty} - \gE \gD \vz^{k}, \vz^{\infty} - \vz^{k} \rangle }{ \lVert \gE \gD \vz^{\infty} - \gE \gD \vz^{k} \rVert_2^2} \geq \beta > 0. 
\end{equation}
Equations~(\ref{eqn:cocoercivity}) and (\ref{eqn:cocoercivity_y}) are equivalent to \cref{eqn:actual_checking} if $\vz^\star = \vz^\infty$ and $\vz^k = \vy^k$.
By \ref{thm2:conclusion:3} and \ref{thm2:conclusion:4} of \Cref{thm:2}, if \cref{eqn:actual_checking} holds while $\vz^k \rightarrow \vz^\star$, then the convergence was due to \cref{eqn:actual_checking}.

\begin{figure}[!h]
    \centering
    \begin{subfigure}[b]{0.32\textwidth}
        \centering
        \includegraphics[width=\textwidth]{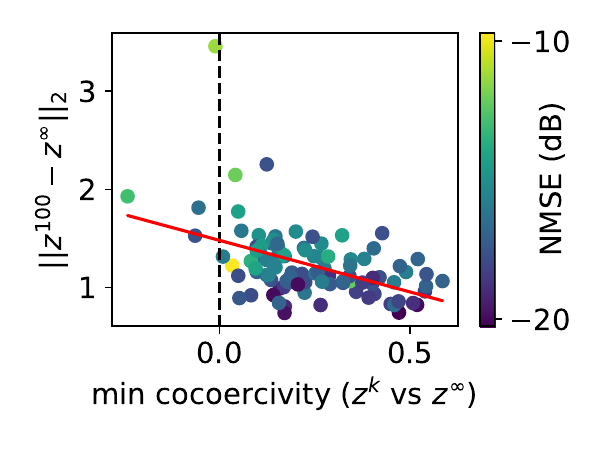}
        \caption{SD 2.1~\citep{rombach2022high}, vanilla FSM}
        \label{fig:sub1}
    \end{subfigure}
    \begin{subfigure}[b]{0.32\textwidth}
        \centering
        \includegraphics[width=\textwidth]{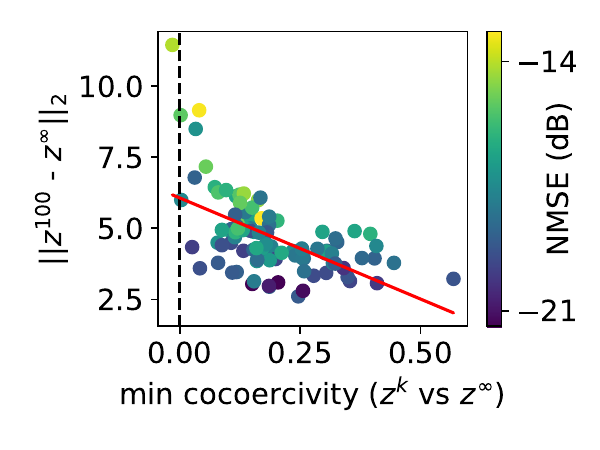}
        \caption{LaVie~\citep{wang2023lavie}, vanilla FSM}
        \label{fig:sub2}
    \end{subfigure}
    \begin{subfigure}[b]{0.32\textwidth}
        \centering
        \includegraphics[width=\textwidth]{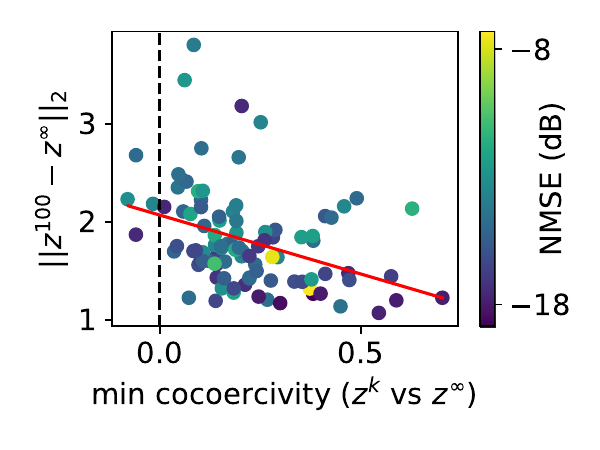}
        \caption{InstaFlow~\citep{liu2023instaflow}, vanilla FSM}
        \label{fig:sub3}
    \end{subfigure}
    \begin{subfigure}[b]{0.32\textwidth}
        \centering
        \includegraphics[width=\textwidth]{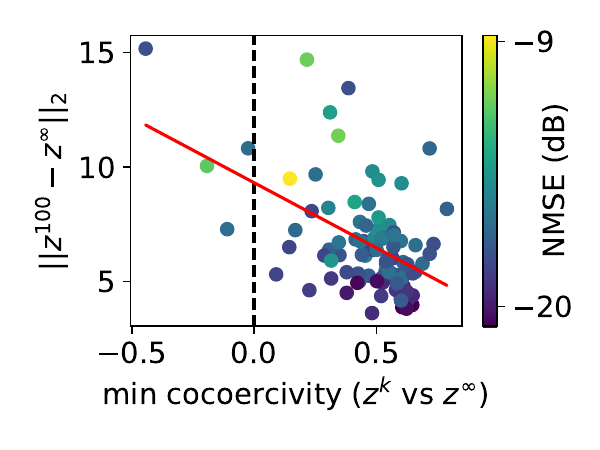}
        \caption{SD 2.1~\citep{rombach2022high}, inertial KM iter.}
        \label{fig:sub4}
    \end{subfigure}
    \begin{subfigure}[b]{0.32\textwidth}
        \centering
        \includegraphics[width=\textwidth]{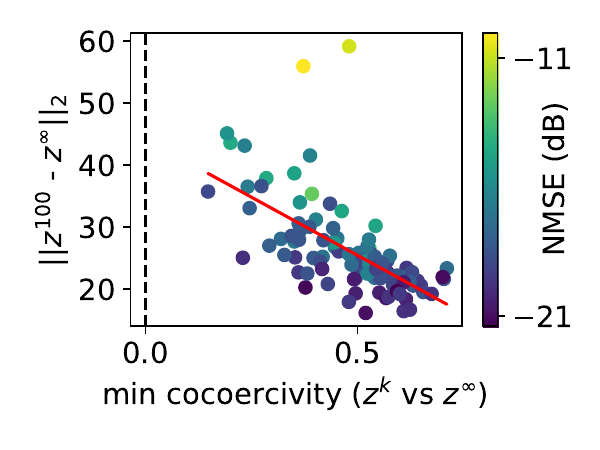}
        \caption{LaVie~\citep{wang2023lavie}, inertial KM iter.}
        \label{fig:sub5}
    \end{subfigure}
    \begin{subfigure}[b]{0.32\textwidth}
        \centering
        \includegraphics[width=\textwidth]{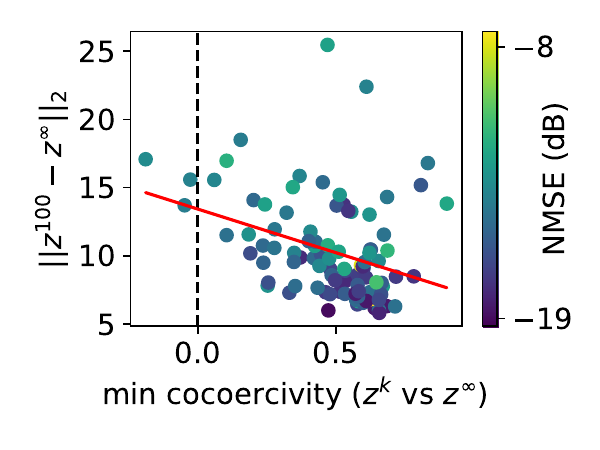}
        \caption{InstaFlow~\citep{liu2023instaflow}, inertial KM iter.}
        \label{fig:sub6}
    \end{subfigure}
.    \caption{The six subfigures represent the relationship between cocoercivity and convergence for 3 models $\times$ 2 algorithms (vanilla forward step method and inertial KM iteration). The x-axis represents the values of $\underset{k\in[0,100]}{\min}\frac{\langle \gE \gD \vz^{\infty} - \gE \gD \vz^{k}, \vz^{\infty} - \vz^{k} \rangle }{ \lVert \gE \gD \vz^{\infty} - \gE \gD \vz^{k} \rVert_2^2}$, which informs whether the optimization path satisfies the assumptions of \Cref{thm:1,thm:2}, while the y-axis represents the convergence (\ie, $\lVert \vz^{100} - \vz^\infty \rVert_2$). The red line shows the linear function fitted by least squares. We set $\vz^{\infty}=\vz^{300}$.}
    \label{fig:cocoercivity}
\end{figure}

We conducted decoder inversion experiments in various recent LDMs such as Stable Diffusion 2.1~\citep{rombach2022high}, LaVie~\citep{wang2023lavie}, and InstaFlow~\citep{liu2023instaflow}. With or without momentum, the learning rate $\rho$ was fixed at 0.001. For inertial KM iterations, $\alpha$ was set to 0.9. \Cref{fig:cocoercivity} illustrates the $\min_k$ of cocoercivity (\ref{eqn:actual_checking}), convergence of the algorithm ($\| \vz^{100} - \vz^\infty \|_2$), and final NMSE ($\| \vz^\infty - \vz^\star \|_2^2 / \|\vz^\star \|_2^2$) for 100 instances in each scenario. The observations and the resulting insights are as follows:
\begin{enumerate}
    \item \textbf{Our assumption on cocoercivity is reasonable:} most of the instances showed $\frac{\langle \gE \gD \vz^{\infty} - \gE \gD \vz^{k}, \vz^{\infty} - \vz^{k} \rangle }{ \lVert \gE \gD \vz^{\infty} - \gE \gD \vz^{k} \rVert_2^2} > 0$ for all $k$. 
    \item \textbf{The better the convergence, the more cocoercive it is:} the fitted functions (red lines) have negative slopes.
    \item \textbf{Converging instances closely approximate the ground truth:} the points at the bottom are darker. 
\end{enumerate}
Those findings support the validity of our assumptions and theorems. 
  
\begin{figure}[!b]
\begin{subfigure}[b]{\textwidth}
    \begin{subfigure}[b]{0.5\textwidth}
    \centering
    \small
    \begin{tikzpicture}
    \begin{axis}[
        xlabel={Runtime for decoder inversion (sec)},
        ylabel={Noise recon. NMSE (dB)},
        legend style={
            at={(1.5,0.7)},
            anchor=south,
            legend columns=1,
            cells={anchor=west},
            rounded corners=2pt,
            draw=none,
        },
        grid=both,
        grid style={dashed,gray!30},
        xmin=0, xmax=36,
        ymin=-25.5, ymax=-14.5,
        xtick={0,10,20,30},
        width=0.9\textwidth,
        height=0.55\textwidth,
    ]

    \addplot [draw=none, fill=black!20, fill opacity=0.8] coordinates {
        (6.1, -15.92)
        (6.29, -18.614)
        (9.371, -19.509)
        (15.8, -20.69)
        (16.046, -20.926)
        (31.7, -24.82)
        (36, -24.82)
        (36, -14.5) 
        (6.1, -14.5) 
    };
    
    \plotblack{    (6.29, -18.614) +- (0, 1.96*0.2484)}; 
    \plotblack{    (9.371, -19.509) +- (0, 1.96*0.2815)}; 
    \plotblack{    (16.046, -20.926) +- (0, 1.96*0.3398)}; 
    \plotblack{    (32.836, -22.998) +- (0, 1.96*0.4319)}; 
    \plotblack{    (6.1, -15.92) +- (0, 1.96*0.225)}; 
    \plotblack{    (9.3, -17.84) +- (0, 1.96*0.28)}; 
    \plotblack{    (15.8, -20.69) +- (0, 1.96*0.37)}; 
    \plotblack{    (31.7, -24.82) +- (0, 1.96*0.55)}; 

    \plotmagenta{ (3.337,-19.392) +- (0,1.96*0.2736) };
    \plotmagenta{ (8.526,-20.835) +- (0,1.96*0.3376) };
    \plotmagenta{ (17.676,-21.706) +- (0,1.96*0.391) };
    \plotmagenta{ (34.473, -21.854) +- (0,1.96*0.4201) };

     \plotcyan{    (1.91,-18.74) +- (0,1.96*0.2448)};
    \plotcyan{   (4.899,-20.824) +- (0,1.96*0.3369)};
     \plotcyan{    (9.56,-21.724) +- (0,1.96*0.3915)};
     \plotcyan{   (19.323, -21.859) +- (0,1.96*0.4185)};


    \end{axis}
    \end{tikzpicture}
    \begin{tikzpicture}[remember picture, overlay]
        \node[draw, minimum size=0.2cm] at (-4.9,3.5) {};
        \node[align=left, font=\sffamily\scriptsize] at (-4.1,3.53) {Grad-based \\32-bit};
        \draw[magenta] (-3.2,3.43) -- ++(60:0.2cm) -- ++(-60:0.2cm) -- cycle;
        \node[align=left, font=\sffamily\scriptsize] at (-2.4,3.53) {Grad-free\\32-bit};
        \filldraw[cyan] (-1.5,3.5) circle (0.1cm);
        \node[align=left, font=\sffamily\scriptsize] at (-0.8,3.53) {Grad-free\\16-bit};
        
        \node[align=left, below right, font=\sffamily\scriptsize] at (-5.1, 1.55) {We seek \\ bottom \& left};
        \node[align=left, below right, font=\sffamily] at (-2, 3.2) {Pareto \\ inefficient };
    \end{tikzpicture}
     \end{subfigure}
    \begin{subfigure}[b]{0.48\textwidth}
    \small
    \begin{bchart}[step=2, max=10, width=0.55\textwidth]
        \bcbar[label=Grad-based, text=32-bit, color=white]{8.83}
        \smallskip
        \bcbar[text=32-bit, color=half_magenta]{3.81}
        \bclabel{Grad-free (ours)}
        \bcbar[text=16-bit, color=half_cyan]{3.15}
        \bcxlabel{Peak memory usage (GB)}
    \end{bchart}
     \end{subfigure}
    \caption{Stable Diffusion 2.1~\citep{rombach2022high}. }\label{fig:main_result_sd2.1}
    
\end{subfigure}
\\
\begin{subfigure}[b]{\textwidth}
    \begin{subfigure}[b]{0.5\textwidth}
    \centering
    \small
    \begin{tikzpicture}
    \begin{axis}[
        xlabel={Runtime for decoder inversion (sec)},
        ylabel={Noise recon. NMSE (dB)},
        legend style={
            at={(1.5,0.7)},
            anchor=south,
            legend columns=1,
            cells={anchor=west},
            rounded corners=2pt,
            draw=none,
        },
        grid=both,
        grid style={dashed,gray!30},
        xmin=0, xmax=130,
        ymin=-23.5, ymax=-16.5,
        xtick={0,40,80,120},
        ytick={-23,-20,-17},
        width=0.9\textwidth,
        height=0.55\textwidth,
    ]

    Shaded area above Pareto front
    \addplot [draw=none, fill=black!20, fill opacity=0.8] coordinates {
        (24.863, -17.352)
        (25.11, -19.06)
        (37.11, -19.68)
        (61.67, -20.66)
        (122.306, -22.201)
        (130, -22.201)
        (130, -16.5) 
        (24.863, -16.5) 
    };
    
    \plotblack{(25.11, -19.06) +- (0, 1.96*0.2213)}; 
    \plotblack{(37.11, -19.68) +- (0, 1.96*0.2451)}; 
    \plotblack{(61.67, -20.66) +- (0, 1.96*0.2885)}; 
    \plotblack{(122.463, -22.140) +- (0, 1.96*0.3600)}; 

    \plotblack{(24.863, -17.352) +- (0, 0.4518)}; 
    \plotblack{(37.072, -18.115) +- (0, 0.4855)}; 
    \plotblack{(61.37, -19.58) +- (0, 0.5941)}; 
    \plotblack{(122.306, -22.201) +- (0, 0.8083)}; 

    \plotmagenta{(12.96,-20.21) +- (0, 0.4306)}; 
    \plotmagenta{(32.167,-21.418) +- (0, 0.5431)};
    \plotmagenta{(64.25,-21.577) +- (0, 0.6229)};
    \plotmagenta{(128.337,-21.211) +- (0, 0.5762)};

    \plotcyan{(8.584,-19.86) +- (0, 0.4212)}; 
    \plotcyan{(21.19,-21.371) +- (0, 0.5459)};
    \plotcyan{(42.281,-21.524) +- (0, 0.6343)};
    \plotcyan{(83.448, -21.135) +- (0, 0.5811)};
    

    \end{axis}
    \end{tikzpicture}
    \begin{tikzpicture}[remember picture, overlay]
        \node[draw, minimum size=0.2cm] at (-4.9,3.5) {};
        \node[align=left, font=\sffamily\scriptsize] at (-4.1,3.53) {Grad-based \\32-bit};
        \draw[magenta] (-3.2,3.43) -- ++(60:0.2cm) -- ++(-60:0.2cm) -- cycle;
        \node[align=left, font=\sffamily\scriptsize] at (-2.4,3.53) {Grad-free\\32-bit};
        \filldraw[cyan] (-1.5,3.5) circle (0.1cm);
        \node[align=left, font=\sffamily\scriptsize] at (-0.8,3.53) {Grad-free\\16-bit};
        
        \node[align=left, below right, font=\sffamily\scriptsize] at (-5.1, 1.55) {We seek \\ bottom \& left};
        \node[align=left, below right, font=\sffamily] at (-2, 3.2) {Pareto \\ inefficient };
    \end{tikzpicture}
    \end{subfigure}
    \begin{subfigure}[b]{0.40\textwidth}
    \small
    \begin{bchart}[step=10, max=70, width=0.70\textwidth]
        \bcbar[label=Grad-based, text=32-bit, color=white]{64.7}
        \smallskip
        \bcbar[text=\scriptsize 32, color=half_magenta]{11.7}
        \bclabel{Grad-free (ours)}
        \bcbar[text=\scriptsize 16, color=half_cyan]{7.13}
        \bcxlabel{Peak memory usage (GB)}
    \end{bchart}

     \end{subfigure}
    \caption{LaVie, a video diffusion model~\citep{wang2023lavie}}\label{fig:main_result_lavie}
\end{subfigure}
\\
\begin{subfigure}[b]{\textwidth}
    \begin{subfigure}[b]{0.5\textwidth}
    \centering
    \small
    \begin{tikzpicture}
    \begin{axis}[
        xlabel={Runtime for decoder inversion (sec)},
        ylabel={Noise recon. NMSE (dB)},
        legend style={
            at={(1.5,0.7)},
            anchor=south,
            legend columns=1,
            cells={anchor=west},
            rounded corners=2pt,
            draw=none,
        },
        grid=both,
        grid style={dashed,gray!30},
        xmin=0, xmax=40,
        ymin=-20, ymax=-15,
        xtick={0,10,20,30,40},
        width=0.9\textwidth,
        height=0.55\textwidth,
    ]
    

    \addplot [draw=none, fill=black!20, fill opacity=0.8] coordinates {
        (6.29, -15.67)
        (6.30, -15.94)
        (9.51, -16.40)
        (16.12, -17.26)
        (32.54, -19.15)
        (40.54, -19.15) 
        (40.00, -15.00) 
        (6.29, -15.00) 
    };
    
    \plotblack{(6.30, -15.94) +- (0, 0.48)}; 
    \plotblack{(9.51, -16.40) +- (0, 0.52)}; 
    \plotblack{(15.92, -17.14) +- (0, 0.59)}; 
    \plotblack{(32.14, -18.33) +- (0, 0.70)}; 

    \plotblack{(6.29, -15.67) +- (0, 0.48)}; 
    \plotblack{(9.55, -16.22) +- (0, 0.53)}; 
    \plotblack{(16.12, -17.26) +- (0, 0.62)}; 
    \plotblack{(32.54, -19.15) +- (0, 0.78)}; 

    \plotmagenta{(3.17,-16.55) +- (0, 0.53)}; 
    \plotmagenta{(8.34,-17.46) +- (0, 0.63)};
    \plotmagenta{(16.90,-17.76) +- (0, 0.68)};
    \plotmagenta{(34.21,-17.81) +- (0, 0.70)};

    \plotcyan{(1.89,-16.38) +- (0, 0.51)}; 
    \plotcyan{(4.93,-17.46) +- (0, 0.63)};
    \plotcyan{(9.99, -17.76) +- (0, 0.68)};
    \plotcyan{(20.04, -17.82) +- (0, 0.71)};

    \draw[->, thick , line width=1.5pt] (axis cs:5,-20.3) --  (axis cs:0.1,-20.9);
    \node[align=left, below right, font=\sffamily] at (axis cs:5, -19.8) {We seek methods \\ bottom \& left};
    \end{axis}
    \end{tikzpicture}
    \begin{tikzpicture}[remember picture, overlay]
        \node[draw, minimum size=0.2cm] at (-4.9,3.5) {};
        \node[align=left, font=\sffamily\scriptsize] at (-4.1,3.53) {Grad-based \\32-bit};
        \draw[magenta] (-3.2,3.43) -- ++(60:0.2cm) -- ++(-60:0.2cm) -- cycle;
        \node[align=left, font=\sffamily\scriptsize] at (-2.4,3.53) {Grad-free\\32-bit};
        \filldraw[cyan] (-1.5,3.5) circle (0.1cm);
        \node[align=left, font=\sffamily\scriptsize] at (-0.8,3.53) {Grad-free\\16-bit};

        \node[align=left, below right, font=\sffamily\scriptsize] at (-5.1, 1.55) {We seek \\ bottom \& left};
        \node[align=left, below right, font=\sffamily] at (-2, 3.2) {Pareto \\ inefficient };
    \end{tikzpicture}
     \end{subfigure}
    \begin{subfigure}[b]{0.48\textwidth}
    \small
    \begin{bchart}[step=2, max=10, width=0.55\textwidth]
        \bcbar[label=Grad-based, text=32-bit, color=white]{8.38}
        \smallskip
        \bcbar[text=32-bit, color=half_magenta]{3.29}
        \bclabel{Grad-free (ours)}
        \bcbar[text=16-bit, color=half_cyan]{2.62}
        \bcxlabel{Peak memory usage (GB)}
    \end{bchart}
     \end{subfigure}
    \caption{InstaFlow, one-step image generating model~\citep{liu2023instaflow}. }\label{fig:main_result_rf}
    
\end{subfigure}
\caption{Our gradient-free decoder inversion has a way shorter runtime than the gradient-based decoder inversion, and drastically reduces the GPU memory usage, on (a) SD2.1~\citep{rombach2022high}, (b) LaVie~\citep{wang2023lavie}, and (c) InstaFlow~\citep{liu2023instaflow}. Note that 16-bit gradient-based approach is unimplementable, due to the underflow problem. Each point represents a different hyperparameter setting (e.g., the total number of iterations, learning rate, learning rate scheduling); by collecting experimental results from such diverse settings, the Pareto frontier can be obtained fairly without manipulation.}\label{fig:main_result}
\end{figure}

\section{Experiments with practical optimization techniques}\label{sec:experiments}

\paragraph{Adam optimizer.} In \cref{sec:method}, we proved that our proposed forward step method (\cref{thm:1}) and inertial KM iterations (\cref{thm:2}) converge. Moreover, we experimentally showed that the assumptions of the theorems hold for most instances and demonstrated their convergence as predicted (\cref{fig:cocoercivity}).
However, it is generally believed that using the Adam optimizer~\citep{kingma2014adam} is preferable in many optimization problems. In this section, although we cannot prove convergence as in \Cref{thm:1,thm:2}, we empirically demonstrate that using the Adam optimizer can achieve a good runtime and memory usage compared to conventional gradient-based methods.

\paragraph{(Optional) Learning rate scheduling.}
Following common beliefs and a prior work which solves the same problem~\citep{hong2023exact}, 
one can also use a cosine learning rate scheduler with warm-up steps. Similar to \citep{hong2023exact}, the first \nicefrac{1}{10} of the steps are warm-up steps, followed by the application of cosine annealing. After \nicefrac{8}{10} of the total steps have passed (as the learning rate has sufficiently decreased), it is kept constant for the rest of steps.


\paragraph{Results.} \Cref{fig:main_result} shows the average NMSE and peak memory usage when performing decoder inversion using practical techniques (\ie, Adam and learning rate scheduling) in three different LDMs. While the gradient-based method required much runtime and GPU memory to achieve a certain accuracy, our approach achieves good accuracy in much less runtime and memory usage. One advantage of our proposed method is that it enables all operations to be performed in 16-bit through gradient-free methods, which would typically be infeasible with gradient-based approaches~\cite{micikevicius2017mixed}. As a result, for video LDMs where large vectors need to be estimated (\cref{fig:main_result_lavie}), memory usages can be significantly reduced by almost 9 times.

\section{Application: Tree-rings watermarking for image generation}\label{sec:app}
In this section, we show an interesting application of our gradient-free decoder inversion.
\citet{wen2023tree} proposed \emph{tree-rings watermarking}, which is an invisible and robust method for protecting the copyright of diffusion-generated images. Watermark is embedded into the Fourier transform of $\vz_T$, and detected by inversion (\ie, estimating $\vz_T$ from $\vx$). \citet{hong2023exact} went beyond watermark detection to attempt watermark \emph{classification}, which makes the problem more difficult; the inversion should be even more accurate.
In the following scenarios with two image generation LDMs, we experimentally show that our decoder inversion can efficiently perform watermark classification.

\paragraph{Watermark classification in SD 2.1.}  Since watermark classification requires higher accuracy than detection, iterative diffusion inversion algorithms was used such as the backward Euler instead of the na\"ive DDIM inversion. 
However, it is reported that iterative algorithms become unstable when the classifier-free guidance is large~\cite{Pan_2023_ICCV,hong2023exact}. So even for watermark classification, the na\"ive DDIM inversion should be used for now. In this case, one promising option for improving the performance of watermark classification is \emph{decoder inversion}.

\paragraph{Watermark classification in InstaFlow.} 
One-step models such as InstaFlow~\citep{liu2023instaflow} generate $\vz_{0}$ directly from $\vz_{T}$. Therefore, rather than the na\"ive DDIM inversion, the backward Euler method\citep{hong2023exact} should be used to obtain $\vz_{T}$ from $\vz_{0}$. In this case, $\vz_{0}$ is very sensitive to $\vz_{T}$, so $\vz_{T}$ should be estimated accurately by \emph{decoder inversion}.

\paragraph{Results.} \Cref{tab:wm_classification} shows the accuracy, peak memory usage and runtime of classifying three tree-rings watermarks using different decoder inversion algorithms.
Our decoder inversion method achieves watermark classification performance comparable to a gradient-based method while significantly reducing runtime and peak memory usage. See the supplementary material for the details.

\begin{table}[!h]
    \centering
    \caption{Our gradient-free decoder inversion method achieves the tree-rings watermark~\citep{wen2023tree} classification performance comparable to a gradient-based method while significantly reducing memory usage and runtime, in two different scenarios.}
    \label{tab:wm_classification}
    \begin{tabular}{l|l|ccc}
    \toprule
         LDM & & Encoder & Gradient-based~\citep{hong2023exact} & Gradient-free (ours) \\ \midrule \hline
    \multirow{3}{*}{SD2.1~\cite{rombach2022high}} & Accuracy     & \cellcolor{tabthird}186/300 & \cellcolor{tabfirst}207/300 & \cellcolor{tabsecond}202/300 \\ \cline{2-5}
    & Peak memory (GB) & \cellcolor{tabfirst}5.71 & \cellcolor{tabthird}11.4 & \cellcolor{tabsecond}6.35 \\ \cline{2-5}
    & Runtime (s) & \cellcolor{tabfirst}5.66 & \cellcolor{tabthird}38.0 & \cellcolor{tabsecond}22.9 \\ 
     \hline \hline
    \multirow{3}{*}{InstaFlow~\cite{liu2023instaflow}} & Accuracy     & \cellcolor{tabthird}149/300 & \cellcolor{tabfirst}227/300 & \cellcolor{tabfirst}227/300 \\ \cline{2-5}
    & Peak memory (GB) & \cellcolor{tabfirst}2.93 & \cellcolor{tabthird}8.84 & \cellcolor{tabsecond}3.15 \\ \cline{2-5}
    & Runtime (s) & \cellcolor{tabfirst}3.55 & \cellcolor{tabthird}35.9 & \cellcolor{tabsecond}13.6 \\ \hline
    \end{tabular}

\end{table}



\section{Discussion}
\subsection{Does the $\beta$-cocoercivity hold also when using Adam? Yes.}
\begin{figure}[h]
  \centering
  \begin{subfigure}{0.32\textwidth}
    \centering
    \includegraphics[width=\textwidth]{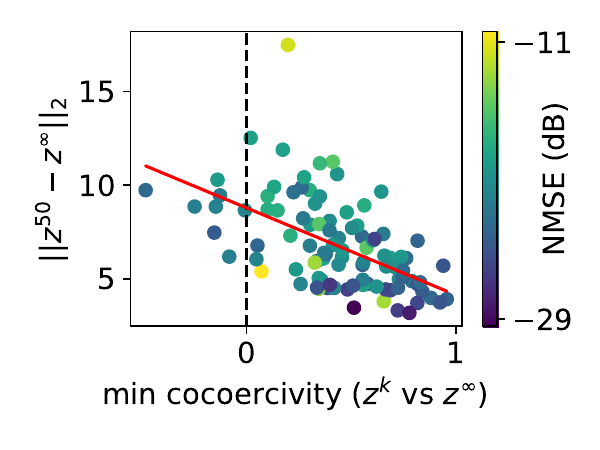} 
    \caption{Stable Diffusion 2.1~\citep{rombach2022high}}
    \label{fig:cocoercivity_adam_1}
  \end{subfigure}
  \begin{subfigure}{0.32\textwidth}
    \centering
    \includegraphics[width=\textwidth]{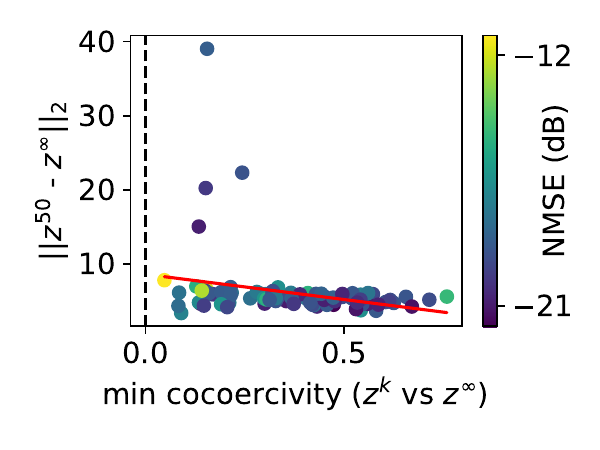} 
    \caption{LaVie~\citep{wang2023lavie}}
    \label{fig:cocoercivity_adam_2}
  \end{subfigure}
  \begin{subfigure}{0.32\textwidth}
    \centering
    \includegraphics[width=\textwidth]{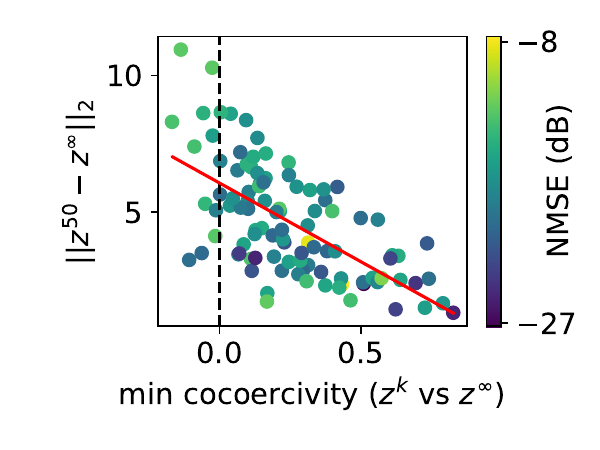} 
    \caption{InstaFlow~\citep{liu2023instaflow}}
    \label{fig:cocoercivity_adam_3}
  \end{subfigure}
  \caption{The assumptions also hold for Adam in many cases, as similar in \cref{fig:cocoercivity}. The x-axis represents the values of $\underset{k\in[0,50]}{\min}\frac{\langle \gE \gD \vz^{\infty} - \gE \gD \vz^{k}, \vz^{\infty} - \vz^{k} \rangle }{ \lVert \gE \gD \vz^{\infty} - \gE \gD \vz^{k} \rVert_2^2}$, which informs whether the optimization path satisfies the assumptions of \Cref{thm:1,thm:2}, while the y-axis represents the convergence (\ie, $\lVert \vz^{50} - \vz^\infty \rVert_2$). The red line shows the linear function fitted by least squares. Note that $\vz^{\infty}$ was approximated by $\vz^{300}$.}
  \label{fig:cocoercivity_adam}
\end{figure}

Although it is challenging to provide a proof regarding convergence for algorithms using Adam and learning rate scheduling (those used in \cref{sec:experiments} and \cref{fig:main_result_rf}), we can rather test if $(\vz^k)$ satisfy cocoercivity (\ref{eqn:cocoercivity}), as in \cref{fig:cocoercivity}.
\Cref{fig:cocoercivity_adam} depicts the plot of \cref{fig:cocoercivity} under the conditions of the experiments in \cref{sec:experiments}, specifically with 16-bit precision and 50 iterations (as Adam converges faster than others). In the cases of \cref{fig:cocoercivity_adam_2,fig:cocoercivity_adam_3}, cocoercivity is well satisfied, and even in \cref{fig:cocoercivity_adam_1}, many instances meet the cocoercivity requirement. Additionally, as in \cref{fig:cocoercivity}, the fitted functions have a negative slope. These findings show that our analysis can help explain some aspects of Adam's convergence.

\subsection{Why this method has not been proposed in GAN inversion studies}
Some may wonder why this gradient-free approach (replacing gradient descent of \cref{eqn:gd} with a forward step method of \cref{eqn:fsm} using the encoder) has not been proposed in numerous GAN inversion studies. Here, we provide \cref{tab:reason} to explain the reasons. The first reason is that GAN inversion cannot use the encoder from off-the-shelf~\citep{xia2022gan}. It is because networks commonly used in GAN inversion, such as StyleGAN~\citep{abdal2019image2stylegan}, were not born with an encoder even though there are some GAN methods that are inherently an autoencoder such as VQGAN~\citep{esser2021taming} that was born with an encoder. Thus many GAN inversion works needed to train the encoder~\citep{perarnau2016invertible, creswell2018inverting, pidhorskyi2020adversarial, zhu2020domain, chai2021using, richardson2021encoding, chai2021ensembling, alaluf2022hyperstyle}, while we can use the encoder without any finetuning. The second reason is that the memory requirement was small in GAN inversion, as the size of the generated images was small. Even a standard GPU on a PC was capable of running gradient-based algorithms, so there was not much need for lighter algorithms than gradient-based methods. On the contrary, due to the increasing size of images/videos generated by recent LDMs, performing decoder inversion with gradient-based algorithms requires huge memory. Another reason is that many networks for GAN inversion operate in full precision (32 bits), making it easier to employ gradient-based methods here. Meanwhile, many recent LDMs often infer in half-precision (16 bits) to enhance efficiency, making gradient-based methods challenging~\citep{micikevicius2017mixed}.

\begin{table}[!h]
    \centering
    \caption{Why has a gradient-free method not been proposed for GAN inversion?}
    \label{tab:reason}
    \centering
    \begin{tabular}{l|cc}
        \toprule
         & GAN inversion & Decoder inversion in LDMs\\ \midrule
         Encoder usage & Limited & Very easy \\
         Memory requirement & Small & Large (\eg, video generation) \\ 
         Inference precision & Full (32-bit) & Half (16-bit) \\ \bottomrule
    \end{tabular}
\end{table}



\section{Conclusion}
In this work, we proposed gradient-free decoder inversion methods for LDMs. The vanilla forward step method and its extension with momentum were shown to guarantee convergence, and the assumptions and theorems were validated for various LDMs. We experimentally showed that a practical algorithm significantly reduced runtime and memory usage compared to existing gradient-based methods. 

\paragraph{Limitations}
Existing gradient-based method is more accurate than our method if sufficient runtime and GPU memory are available. In applications such as image editing, where the accuracy of latent reconstruction may not be critically important such as background preservation, decoder inversion is often unnecessary in most settings (\ie, using the encoder is just OK).

\paragraph{Broader impacts}
This study could positively impact the protection of copyright for LDM's outputs. One day, even if a new neural network architecture is invented and diffusion is no longer used, our method can still be efficiently utilized to improve the accuracy of decoder inversion as long as transformations between latent and pixel-space are required.

\section*{Acknowledgements}
This work was supported by Institute of Information \& communications Technology Planning \& Evaluation (IITP) grant funded by the Korea government(MSIT) [NO.2021-0-01343, Artificial Intelligence Graduate School Program (Seoul National University)] and National Research Foundation of Korea (NRF) grants funded by the Korea government (MSIT) (NRF-2022R1A4A1030579, NRF-2022M3C1A309202211). The authors acknowledged the financial support from the BK21 FOUR program of the Education and Research Program for Future ICT Pioneers, Seoul National University.

{
    \small
    \bibliographystyle{ieeenat_fullname}
    \bibliography{refs}
}

\appendix

\renewcommand{\thefigure}{S\arabic{figure}}
\renewcommand{\thetable}{S\arabic{table}}
\renewcommand{\theequation}{S\arabic{equation}}
\renewcommand{\thetheorem}{S\arabic{theorem}}
\renewcommand{\thelemma}{S\arabic{lemma}}
\renewcommand{\theassumption}{S\arabic{assumption}}

\setcounter{figure}{0}
\setcounter{table}{0}
\setcounter{equation}{0}
\setcounter{theorem}{0}
\setcounter{lemma}{0}
\setcounter{assumption}{0}

\section{Appendix / supplemental material}
\subsection{Proofs}
Throughout the paper and supplementary material, both the inner product and the norm are $l_2$.
\subsubsection{Proof of \cref{thm:1}}
\begin{proof}[Proof of \cref{thm:1}]
    \begin{equation}
    \begin{aligned}
        \lVert \vz^{k+1} - \vz^\star \rVert_2^2 & = \lVert \vz^{k} - \vz^\star - \rho \gT \vz^k \rVert_2^2 \\
        & = \lVert \vz^{k} - \vz^\star \rVert_2^2 - 2\rho \langle \gT \vz^k, \vz^k - \vz^\star \rangle + \rho^2 \lVert \gT \vz^k \rVert_2^2\\
        & \leq \lVert \vz^{k} - \vz^\star \rVert_2^2 - \rho (2\beta - \rho) \lVert \gT \vz^k \rVert_2^2\\
    \end{aligned}
    \end{equation}
    Sum both sides from $k=0, \dots, \infty$
    \begin{equation}
        \rho(2\beta - \rho) \sum_{k=0}^\infty \lVert \gT \vz^k \rVert_2^2 \leq \lVert \vz^0 - \vz^\star \rVert_2^2 < \infty \Rightarrow \lVert \gT \vz^k \rVert_2^2 \rightarrow 0 \Rightarrow \gT\vz^k \rightarrow 0. 
    \end{equation}
    Finally, if $\vz^k \rightarrow \vz^\infty$, then $\gT\vz^\infty=0$ by continuity.
\end{proof}

\subsubsection{Lemma 1 for Theorem 2}
For simplicity, we set
\begin{align}
    & \nu = (\lambda^{-1} - 1),\label{eqn:label:nu}\\
    & \delta^k = \nu (1-\alpha) \| \vz^k - \vz^{k-1} \|^2,\label{eqn:label:delta}\\
    & \Delta^k(\vz^\star)= \|\vz^k - \vz^\star \|^2 - \|\vz^{k-1} - \vz^\star \|^2, \quad \Delta^1(\vz^\star) = 0 \label{eqn:label:Delta}\\
    & C^k (\vz^\star) = \| \vz^k - \vz^\star \|^2 - \alpha \| \vz^{k-1} - \vz^\star \|^2 + \delta^k, \quad C^1 (\vz^\star) = \| \vz^1 - \vz^\star \|^2.\label{eqn:label:Ck}
\end{align}

Before we prove, we need the following lemma.
\begin{lemma}\label{lemma1}
    Let $\beta>0$, $0<\rho<2\beta$, and $\vx \in \R^N$. Assume $\gT(\cdot)=\gE \circ \gD(\cdot) - \gE(\vx)$ is continuous. Assume $\gT(\cdot)=\gE \circ \gD(\cdot) - \gE(\vx)$ is continuous. Let $(\vz^k, \vy^k)$ satisfy (\ref{eqn:momentum_1}) and (\ref{eqn:momentum_2}). 
    Assume $\vz^\star$ is a zero of $\gT$ (\ie, $\gT \vz^\star = 0$) and \cref{eqn:cocoercivity_y} holds.
    Then, for $k=1,2,\dots$,
    \begin{equation}\label{eqn:L1:0}
        \Delta^{k+1}(\vz^\star) + \delta^{k+1} + \nu \alpha \| \vz^{k+1} - 2\vz^k + \vz^{k-1}\|^2 \leq \alpha \Delta^k (\vz^\star) + [\alpha(1+\alpha) + \nu \alpha(1-\alpha)] \| \vz^k - \vz^{k-1} \|^2.
    \end{equation}
\end{lemma}
\begin{proof}[Proof of \Cref{lemma1}]
Let $\rho = 2 \lambda \beta$.
From \cref{eqn:momentum_2},
\begin{equation}\label{eqn:L1:1}
        \|\vz^{k+1} - \vz^\star \|^2 = \|\vy^k - \vz^\star \|^2 + \rho^2 \|\gT\vy^k\|^2 - 2\rho \langle \vy^k - \vz^\star, \gT\vy^k \rangle.
\end{equation}
Applying \cref{eqn:cocoercivity_y} on \cref{eqn:L1:1}, we have
\begin{equation}\label{eqn:L1:2}
    \|\vz^{k+1} - \vz^\star \|^2 \leq \|\vy^k - \vz^\star \|^2 - \rho (2\beta - \rho)  \|\gT\vy^k\|^2.
\end{equation}
From \cref{eqn:momentum_1},
\begin{equation}\label{eqn:L1:3}
\begin{aligned}
    \|\vy^k - \vz^\star \|^2 & = \| (1+\alpha)(\vz^k - \vz^\star) - \alpha (\vz^{k-1}-\vz^\star) \|^2\\
    & = (1+\alpha) \| \vz^k - \vz^\star \|^2 + \alpha (1 + \alpha) \|\vz^k - \vz^{k-1}\|^2 - \alpha \|\vz^{k-1}-\vz^\star\|^2.
\end{aligned}
\end{equation}
Combining \cref{eqn:L1:2} and \cref{eqn:L1:3},
\begin{equation}\label{eqn:L1:4}
    \|\vz^{k+1}-\vz^\star \| \leq (1+\alpha) \| \vz^k - \vz^\star \|^2 + \alpha (1 + \alpha) \|\vz^k - \vz^{k-1}\|^2 - \alpha \|\vz^{k-1}-\vz^\star\|^2 - \rho (2\beta - \rho)  \|\gT\vy^k\|^2.
\end{equation}
Using \cref{eqn:label:delta}, we can simplify \cref{eqn:L1:4} as follows:
\begin{equation}\label{eqn:L1:5}
   \Delta^{k+1}(\vz^\star) \leq \alpha \Delta^k (\vz^\star) + \alpha (1+\alpha) \| \vz^k - \vz^{k-1} \|^2 -\rho(2\beta - \rho) \| \gT\vy^k \|^2.
\end{equation}
On the other hand, by \cref{eqn:momentum_1} and \cref{eqn:momentum_2}, we have
\begin{equation}\label{eqn:L1:6}
\begin{aligned}
    \rho^2 \| \gT\vy^k \|^2  & = \| \vz^{k+1}- \vy^k \|^2 \\
    & = \| \vz^{k+1}- \vz^{k} - \alpha(\vz^k- \vz^{k-1}) \|^2 \\
    & =   \| (1-\alpha)(\vz^{k+1} - \vz^k) + \alpha (\vz^{k+1} - 2\vz^k + \vz^{k-1}) \|^2\\
    & = (1-\alpha) \| \vz^{k+1} - \vz^k \|^2 - \alpha (1 - \alpha) \| \vz^k - \vz^{k-1} \|^2 + \alpha \| \vz^{k+1} - 2\vz^k + \vz^{k-1}\|^2.
\end{aligned}
\end{equation}
Let's multiply $\nu = (2\beta - \rho)/\rho$ to \cref{eqn:L1:6}. Then
\begin{equation}\label{eqn:L1:7}
    \rho (2\beta - \rho) \| \gT\vy^k \|^2  = \delta^{k+1} - \nu \alpha (1 - \alpha) \| \vz^k - \vz^{k-1} \|^2 + \nu \alpha \| \vz^{k+1} - 2\vz^k + \vz^{k-1}\|^2
\end{equation}
Finally, subtracting \cref{eqn:L1:7} from \cref{eqn:L1:5}, we obtain \cref{eqn:L1:0}.
\end{proof}

\subsubsection{Proof of \cref{thm:2}}
\begin{proof}[Proof of \cref{thm:2}]
We firstly show that $\left(C^k(\vz^\star)\right)_k$ is nonincreasing and nonnegative, thus \(  \displaystyle \lim_{k\rightarrow\infty}C^k(\vz^\star) \) exists.
By \cref{eqn:T2:0}, there exists $\eps > 0$ such that
\begin{equation}\label{eqn:T2:1}
    \alpha(1+\alpha) + \nu \alpha(1-\alpha) \leq \nu (1-\alpha)-\epsilon.
\end{equation}
Combining \cref{eqn:L1:0} and \cref{eqn:T2:1}, we have
\begin{equation}\label{eqn:T2:2}
\begin{aligned}
    & \Delta^{k+1}(\vz^\star) + \delta^{k+1} + \nu \alpha \| \vz^{k+1} - 2\vz^k + \vz^{k-1}\|^2\\ & \quad \leq \alpha \Delta^k (\vz^\star) + [\alpha(1+\alpha) + \nu \alpha(1-\alpha)] \| \vz^k - \vz^{k-1} \|^2\\
    & \quad \leq \alpha \Delta^k (\vz^\star) + [\nu (1-\alpha)-\epsilon] \| \vz^k - \vz^{k-1} \|^2\\
    & \quad = \alpha \Delta^k (\vz^\star) + \delta^k -\epsilon \| \vz^k - \vz^{k-1} \|^2.
\end{aligned}
\end{equation}
From \cref{eqn:label:Ck} and \cref{eqn:label:Delta},
\begin{equation}\label{eqn:T2:3}
\begin{aligned}
    C^{k+1}(\vz^\star) - C^{k}(\vz^\star) & = \Delta^{k+1}(\vz^\star) - \alpha(\|\vz^k - \vz^\star \|^2 - \|\vz^{k-1} - \vz^\star \|^2) + \delta^{k+1} - \delta^k \\
    & = \Delta^{k+1}(\vz^\star)+ \delta^{k+1}  - \alpha\Delta^k(\vz^\star)  - \delta^k. 
\end{aligned}
\end{equation}
By \cref{eqn:T2:2} and \cref{eqn:T2:3}, we have
\begin{equation}\label{eqn:T2:4}
    C^{k+1}(\vz^\star) + \nu \alpha \| \vz^{k+1} - 2\vz^k + \vz^{k-1}\|^2 +\epsilon \| \vz^k - \vz^{k-1} \|^2 \leq C^{k}(\vz^\star),
\end{equation}
which shows $\left(C^k(\vz^\star)\right)_k$ is nonincreasing. To prove it is nonnegative, suppose $C^{k_1}(\vz^\star)<0$ for some $k_1 \geq 1$. Since it is nonincreasing, from \cref{eqn:label:Ck},
\begin{equation}\label{eqn:T2:5}
    \| \vz^k - \vz^\star \|^2 - \alpha \| \vz^{k-1} - \vz^\star \|^2 \leq C^k (\vz^\star) \leq C^{k_1} (\vz^\star) < 0
\end{equation}
for all $k \geq k_1$. That is, $ \| \vz^k - \vz^\star \|^2 \leq \| \vz^{k-1} - \vz^\star \|^2 +   C^{k_1} (\vz^\star)$, hence
\begin{equation}\label{eqn:T2:6}
    0 \leq \| \vz^k - \vz^\star \|^2 \leq \| \vz^{k-1} - \vz^\star \|^2 +   C^{k_1} (\vz^\star) \leq \dots \leq \| \vz^{k_1} - \vz^\star \|^2 + (k-k_1)C^{k_1} (\vz^\star)
\end{equation}
for all $k\geq k_1$, and this is a contradiction as the right hand of \cref{eqn:T2:6} diverges to $-\infty$ as $k \rightarrow \infty$. Therefore, $\left(C^k(\vz^\star)\right)_k$ is nonincreasing and nonnegative, thus \(  \displaystyle \lim_{k\rightarrow\infty}C^k(\vz^\star) \) exists.

Now we begin showing \ref{thm2:conclusion:1}.
By the findings so far and \cref{eqn:T2:4} and since $\epsilon>0$, $\sum_{k\geq 1} \| \vz^{k+1} - 2\vz^k + \vz^{k-1}\|^2$ and  $\sum_{k\geq 1} \| \vz^k - \vz^{k-1} \|^2$ converge. Since $\sum_{k\geq 1} \| \vz^k - \vz^{k-1} \|^2$ converges, $\sum_{k\geq 1} \delta^k = \sum_{k\geq 1} \nu (1-\alpha) \| \vz^k - \vz^{k-1} \|^2 $ also converges.
Using \cref{eqn:L1:6} and an identity $\|a-b\|^2 \leq 2\|a\|^2 + 2\|b\|^2$, we have
\begin{equation}\label{eqn:T2:7}
    \rho^2 \| \gT\vy^k \|^2  = (1+\alpha) \| \vz^{k+1} - \vz^k \|^2 + \alpha (1 + \alpha) \| \vz^k - \vz^{k-1} \|^2.
\end{equation}
Summing \cref{eqn:T2:4} over $k\geq 1$, we have
\begin{equation}\label{eqn:T2:8}
    \epsilon \sum_{k \geq 1} \| \vz^k - \vz^{k-1} \|^2 \leq C^1 (\vz^\star) = \| \vz^1 - \vz^\star \|^2.
\end{equation}
Using the summation of \cref{eqn:T2:7} and \cref{eqn:T2:8}, we have
\begin{equation}\label{eqn:T2:9}
    n \underset{1 \leq k \leq n}{\min} \|\gT\vy^k\|^2 \leq \sum_{k \geq 1} \| \gT\vy^k \|^2 \leq \frac{(1+\alpha)^2}{\epsilon \rho^2} \| \vz^1 - \vz^\star \|^2,
\end{equation}
which shows that $\sum_{k \geq 1} \|\gT \vy^{k} \|^2$ converges, and proves \ref{thm2:conclusion:2}.

Now we prove \ref{thm2:conclusion:3}. From \cref{eqn:T2:2}, we have
\begin{equation}\label{eqn:T2:10}
    \Delta^{k+1}(\vz^\star) \leq \alpha \Delta^{k}(\vz^\star) + \delta^k.
\end{equation}
Let $[\cdot]_+$ denotes the positive part. From \cref{eqn:T2:10},
\begin{equation}\label{eqn:T2:11}
    (1-\alpha)[\Delta^{k+1}(\vz^\star)]_+ + \alpha[\Delta^{k+1}(\vz^\star)]_+ \leq \alpha [\Delta^{k}(\vz^\star)]_+ + \delta^k.
\end{equation}
Applying summation for $k \geq 1$, we have
\begin{equation}\label{eqn:T2:12}
    (1-\alpha)\sum_{k \geq 1}[\Delta^{k+1}(\vz^\star)]_+ \leq \alpha [\Delta^{1}(\vz^\star)]_+ + \sum_{k \geq 1}\delta^k = \sum_{k \geq 1}\delta^k < \infty.    
\end{equation}
Let $h^k = \| \vz^k - \vz^\star \|^2 - \sum_{j=1}^k [\Delta^j(\vz^\star)]_+$. Then $h^{k+1} - h^{k} = \Delta^{k+1}(\vz^\star) - [\Delta^{k+1}(\vz^\star) ]_+ \leq 0$ so $(h^{k})$ is nonincreasing. Since $\sum_{k \geq 1}[\Delta^{k+1}(\vz^\star)]_+$ is finite,  $\lim_{k\rightarrow\infty} \| \vz^k - \vz^\star \| = \lim_{k\rightarrow\infty} h^k$ exists.

Finally, we now prove \ref{thm2:conclusion:4}. By \ref{thm2:conclusion:1}, $\lim_{k\rightarrow \infty} \| \gT \vy^k \| = \lim_{k\rightarrow \infty} \| \vz^k - \vz^{k-1} \| = 0$. By \cref{eqn:momentum_1} and (\ref{eqn:momentum_2}), $(\vy^k)$ and $(\vz^k)$ have the same limit points.
\end{proof}

\begin{table}[!t]
\centering
\caption{We can further improve the reconstruction quality, by using a cosine learning rate scheduler with warm-up steps (if the number of iterations or the runtime is predetermined). We conducted experiments of the decoder inversion in various LDMs, with all other conditions being the same to that in \cref{fig:main_result}, but using a relatively scheduled learning rate at each iteration. $\pm$ represents the 95\% confidence interval.}\label{tab:S1}
\small

\begin{subtable}{\linewidth}
\centering
\caption{Stable Diffusion 2.1}\label{tab:S1a}
\begin{tabular}{@{}c@{ }c@{ }c|cccc|c@{}}
\toprule
Method & Bits & LR  & \multicolumn{4}{c|}{NMSE (dB) // \# iter. / Runtime}        & Memory\\ \midrule
\multirow{2}{*}{Grad-based} & \multirow{2}{*}{32} 
                        &  0.1      & -15.92 \scriptsize{$\pm$ 0.44}  &  -17.84 \scriptsize{$\pm$ 0.55}     &  -20.69 \scriptsize{$\pm$ 0.73}  &  -24.82 \scriptsize{$\pm$ 1.08}              & \multirow{2}{*}{8.83 GB}      \\ \cmidrule{4-7}
                      &  &  scheduled                       & 20 / 6.11 s         & 30 / 9.31 s        & 50 / 15.7 s  & 100\footnote{This setting is the same as the setting in \cite{hong2023exact}.} / 31.7 s       &                                 \\ \midrule
\multirow{2}{*}{Grad-based} & \multirow{2}{*}{32} 
                        &  0.01      & -18.61 \scriptsize{$\pm$ 0.49}  &  -19.51 \scriptsize{$\pm$ 0.55}     &  -20.93 \scriptsize{$\pm$ 0.67}  &  -23.00 \scriptsize{$\pm$ 0.85}              & \multirow{2}{*}{8.83 GB}      \\ \cmidrule{4-7}
                      &  &  fixed                       & 20 / 6.29 s         & 30 / 9.37 s        & 50 / 16.0 s  & 100 / 32.8 s       &                                 \\ \midrule
\multirow{2}{*}{\begin{tabular}{@{}c@{}}
   Grad-free \\ (ours)
\end{tabular}} & \multirow{2}{*}{32} 
                        &   0.01    & -19.39 \scriptsize{$\pm$ 0.54}  &  -20.84 \scriptsize{$\pm$ 0.66}     &  -21.71 \scriptsize{$\pm$ 0.77}  &  -21.85 \scriptsize{$\pm$ 0.82}              & \multirow{2}{*}{3.81 GB}      \\ \cmidrule{4-7}
                      &  &  scheduled                         & 20 / 3.34 s         & 50 / 8.53 s        & 100 / 17.7 s  & 200 / 34.5 s       &                                 \\ \midrule
\multirow{2}{*}{\begin{tabular}{@{}c@{}}
   Grad-free \\ (ours)
\end{tabular}} & \multirow{2}{*}{16} 
                        &   0.01      & -18.74 \scriptsize{$\pm$ 0.48}  &  -20.82 \scriptsize{$\pm$ 0.66}     &  -21.72 \scriptsize{$\pm$ 0.77}  &  -21.86 \scriptsize{$\pm$ 0.82}              & \multirow{2}{*}{3.15 GB}      \\ \cmidrule{4-7}
                      &  &     scheduled                    & 20 / 1.91 s         & 50 / 4.90 s        & 100 / 9.56 s  & 200 / 19.3 s       &                                 \\                    
                      \bottomrule
\end{tabular}
\end{subtable}
\begin{subtable}{\linewidth}
\centering
\caption{LaVie}\label{tab:S1b}
\begin{tabular}{@{}c@{ }c@{ }c|cccc|c@{}}
\toprule
Method & Bits & LR  & \multicolumn{4}{c|}{NMSE (dB) // \# iter. / Runtime}        & Memory\\ \midrule
\multirow{2}{*}{Grad-based} & \multirow{2}{*}{32} 
                        &  0.1      & -17.35 \scriptsize{$\pm$ 0.45}  &  -18.12 \scriptsize{$\pm$ 0.49}     &  -19.58 \scriptsize{$\pm$ 0.59}  &  -22.20 \scriptsize{$\pm$ 0.81}              & \multirow{2}{*}{64.7 GB}      \\ \cmidrule{4-7}
                      &  &  scheduled                       & 20 / 24.86 s         & 30 / 37.07 s        & 50 / 61.37 s  & 100 / 122.31 s       &                                 \\ \midrule
\multirow{2}{*}{Grad-based} & \multirow{2}{*}{32} 
                        &  0.01      & -19.06 \scriptsize{$\pm$ 0.43}  &  -19.68 \scriptsize{$\pm$ 0.48}     &  -20.66 \scriptsize{$\pm$ 0.57}  &  -22.14 \scriptsize{$\pm$ 0.71}              & \multirow{2}{*}{64.7 GB}      \\ \cmidrule{4-7}
                      &  &  fixed                       & 20 / 25.11 s         & 30 / 37.11 s        & 50 / 61.67 s  & 100 / 122.46 s       &                                 \\ \midrule
\multirow{2}{*}{\begin{tabular}{@{}c@{}}
   Grad-free \\ (ours)
\end{tabular}} & \multirow{2}{*}{32} 
                        &   0.01    & -20.21 \scriptsize{$\pm$ 0.43}  &  -21.42 \scriptsize{$\pm$ 0.54}     &  -21.58 \scriptsize{$\pm$ 0.62}  &  -21.21 \scriptsize{$\pm$ 0.58}              & \multirow{2}{*}{11.7 GB}      \\ \cmidrule{4-7}
                      &  &  scheduled                         & 20 / 12.96 s         & 50 / 32.17 s        & 100 / 64.25 s  & 200 / 128.34 s       &                                 \\ \midrule
\multirow{2}{*}{\begin{tabular}{@{}c@{}}
   Grad-free \\ (ours)
\end{tabular}} & \multirow{2}{*}{16} 
                        &   0.01      & -19.86 \scriptsize{$\pm$ 0.42}  &  -21.37 \scriptsize{$\pm$ 0.55}     &  -21.52 \scriptsize{$\pm$ 0.63}  &  -21.14 \scriptsize{$\pm$ 0.58}              & \multirow{2}{*}{7.13 GB}      \\ \cmidrule{4-7}
                      &  &     scheduled                    & 20 / 8.59 s         & 50 / 21.19 s        & 100 / 42.28 s  & 200 / 83.45 s       &                                 \\                    
                      \bottomrule
\end{tabular}
\end{subtable}

\begin{subtable}{\linewidth}
\centering
\caption{InstaFlow}\label{tab:S1c}
\begin{tabular}{@{}c@{ }c@{ }c|cccc|c@{}}
\toprule
Method & bits &      LR        & \multicolumn{4}{c|}{NMSE (dB) // \# iter. / Runtime}      & Memory\\ \midrule
\multirow{2}{*}{Grad-based} & \multirow{2}{*}{32} 
                        & 0.1         & -15.67 \scriptsize{$\pm$ 0.48}  &  -16.22 \scriptsize{$\pm$ 0.53}     &  -17.26 \scriptsize{$\pm$ 0.62}  &  -19.15 \scriptsize{$\pm$ 0.78}              & \multirow{2}{*}{8.38 GB}      \\ \cmidrule{4-7}
                      &  & scheduled                           & 20 / 6.29 s         & 30 / 9.55 s        & 50 / 16.12 s  & 100 / 32.54 s       &                                 \\ \midrule
\multirow{2}{*}{Grad-based} & \multirow{2}{*}{32} 
                        & 0.01         & -15.94 \scriptsize{$\pm$ 0.48}  &  -16.40 \scriptsize{$\pm$ 0.52}     &  -17.14 \scriptsize{$\pm$ 0.59}  &  -18.33 \scriptsize{$\pm$ 0.70}              & \multirow{2}{*}{8.38 GB}      \\ \cmidrule{4-7}
                      &  & fixed                           & 20 / 6.30 s         & 30 / 9.51 s        & 50 / 15.92 s  & 100 / 32.14 s       &                                 \\ \midrule

\multirow{2}{*}{\begin{tabular}{@{}c@{}}
   Grad-free \\ (ours)
\end{tabular}} & \multirow{2}{*}{32} 
                        & 0.01         & -16.55 \scriptsize{$\pm$ 0.53}  &  -17.46 \scriptsize{$\pm$ 0.63}     &  -17.76 \scriptsize{$\pm$ 0.68}  &  -17.81 \scriptsize{$\pm$ 0.70}              & \multirow{2}{*}{3.2 GB}      \\ \cmidrule{4-7}
                      &  & scheduled                           & 20 / 3.17 s         & 50 / 8.34 s        & 100 / 16.90 s  & 200 / 34.21 s       &                                 \\ \midrule
\multirow{2}{*}{\begin{tabular}{@{}c@{}}
   Grad-free \\ (ours)
\end{tabular}} & \multirow{2}{*}{16} 
                        & 0.01         & -16.38 \scriptsize{$\pm$ 0.51}  &  -17.46 \scriptsize{$\pm$ 0.63}     &  -17.76 \scriptsize{$\pm$ 0.68}  &  -17.82 \scriptsize{$\pm$ 0.71}              & \multirow{2}{*}{2.62 GB}      \\ \cmidrule{4-7}
                      &  & scheduled                           & 20 / 1.89 s         & 50 / 4.93 s        & 100 / 9.99 s  & 200 / 20.04 s       &                                 \\                    
                      \bottomrule
\end{tabular}
\end{subtable}
\end{table}

\subsection{Experiment details and more results}

\subsubsection{Decoder inversion in LDMs}
\citet{hong2023exact} successfully employed the gradient-based method in their experiments and completed hyperparameter tuning. Their tailored method may work well in their setting, so we used the code from \citet{hong2023exact}'s official repository\footnote{https://github.com/smhongok/inv-DM} to perform decoder inversion on 100 images generated under the same settings (prompt, classifier-free guidance, random seed). This makes our comparison fair. \Cref{tab:S1} shows the settings, NMSE, number of iterations, runtime, and memory usage of the experiments. Although all the values in \cref{tab:S1} are actually displayed in \cref{fig:main_result}, \Cref{tab:S1} provides additional information, including on which specific settings correspond to the points in \cref{fig:main_result}.

\paragraph{Gradient-based method}
For the gradient-based method (for comparision), we used Adam~\citep{kingma2014adam} with $l_2$-loss. When optionally using a cosine learning rate scheduler, 1/10 of the total steps were warm-up steps (\ie, the learning rate increased linearly). Note that this setting is the same to \citep{hong2023exact}. In \citep{hong2023exact}, the learning rate was 0.1 with 100 iterations, but it showed long runtimes. Thus, we set the iterations to 20, 30, 50, and 100.  
We could observe that reducing the number of iterations decreases the effectiveness of learning rate scheduling. Therefore, we additionally conducted experiments with a fixed learning rate of 0.01, and found that it performed better with a smaller number of iterations. As shown in \cref{tab:S1a,tab:S1c}, when comparing the `0.1 scheduled' and `0.01 fixed' rows under `Grad-based 32-bit', the fixed learning rate outperforms at 20 and 30 iterations, whereas the scheduled learning rate outperforms at 50 and 100 iterations. By comparing existing methods under various hyperparameter settings, we can more reliably establish the Pareto frontier of these methods, ensuring the fairness of our comparisons.

\paragraph{Other details}
For Stable Diffusion 2.1 and InstaFlow, we used prompts from \url{https://huggingface.co/datasets/Gustavosta/Stable-Diffusion-Prompts}. For LaVie, prompts were generated by ChatGPT, given the examples used in \citep{wang2023lavie}. The classifier-free guidance was 3.0 in Stable Diffusion 2.1, 7.5 in LaVie, and 1.0 in InstaFlow. 
For LaVie, we found that 200 iterations for the gradient-free method is too long, so we adjusted the learning rate scheduling as if there were only 100 iterations (fixing the learning rate after 100 iterations).

\subsubsection{Watermark detection}
For both scenarios, the number of steps was 100 and the learning rate was scheduled for both gradient-based and gradient-free methods. The maximum learning rate was 0.1 for grad-based methods and 0.01 for Grad-free methods, which is consistent with the experiments in Section 4. All experiments were conducted in 32-bit. 

For Stable Diffusion 2.1, through decoder inversion, we obtain $\vz_0$ from $\vx$, and perform na\"ive DDIM inversion for 50 steps to obtain $\vz_T$ from $\vz_0$. Except for employing the na\"ive DDIM inversion due to the use of a large classifier-free guidance of 7.5 and resulting stability issues, the experimental settings are the same as those of \citep{hong2023exact}. 
For InstaFlow, the backward Euler (Algorithm 1 of \citep{hong2023exact}) with 100 iterations is applied for seeking $\vz_T$ from $\vz_0$.

\Cref{fig:confusions} shows the confusion matrices for the experiments of \cref{tab:wm_classification} in the main paper. \textcolor{black}{In \Cref{fig:watermarks2}, we also present the qualitative results of applying our algorithm to the watermark classification experiment. Our grad-free method enables precise reconstruction, while reducing the runtime compared to the grad-based method.
}

\begin{figure*}[h]
    \small
    \centering   
    \begin{subfigure}[]{\textwidth}
    \begin{subfigure}[]{0.3\textwidth}
    \centering
        \begin{tikzpicture}
            \begin{axis}[
                    width=0.9\linewidth,  
                    height=0.625\linewidth,  
                    colormap={custom}{
                        color(0cm)=(white);
                        color(1cm)=(yellow);
                        color(2cm)=(green);
                        color(3cm)=(cyan);
                    },
                    xlabel=Predicted,
                    xlabel style={yshift=0pt},
                    ylabel=Actual,
                    ylabel style={yshift=3pt},
                    xticklabels={ 1,  2,  3},
                    xtick={0,...,2},
                    xtick style={draw=none},
                    yticklabels={ 1,  2,  3},
                    ytick={0,...,2},
                    ytick style={draw=none},
                    enlargelimits=false,
                    xticklabel style={},
                    nodes near coords={\pgfmathprintnumber\pgfplotspointmeta},
                    nodes near coords style={
                        yshift=-7pt
                    },
                    point meta min=0,  
                    point meta max=100  
                ]
                \addplot[
                    matrix plot,
                    mesh/cols=3,
                    point meta=explicit,draw=gray
                ] table [meta=C] {
                    x y C
                    0 0 54
                    1 0 14
                    2 0 32

                    0 1 27
                    1 1 47
                    2 1 26

                    0 2 12
                    1 2 3
                    2 2 85

                };
            \end{axis}
        \end{tikzpicture}
    \caption*{Encoder}
    \label{fig:confusion_naive}
    \end{subfigure}
    \hfill
    \begin{subfigure}[]{0.3\textwidth}
        \centering
        \begin{tikzpicture}
            \begin{axis}[
                    width=0.9\linewidth,  
                    height=0.625\linewidth,  
                    colormap={custom}{
                        color(0cm)=(white);
                        color(1cm)=(yellow);
                        color(2cm)=(green);
                        color(3cm)=(cyan);
                    },
                    xlabel=Predicted,
                    xlabel style={yshift=0pt},
                    ylabel=Actual,
                    ylabel style={yshift=3pt},
                    xticklabels={ 1,  2,  3},
                    xtick={0,...,2},
                    xtick style={draw=none},
                    yticklabels={ 1,  2,  3},
                    ytick={0,...,2},
                    ytick style={draw=none},
                    enlargelimits=false,
                    xticklabel style={},
                    nodes near coords={\pgfmathprintnumber\pgfplotspointmeta},
                    nodes near coords style={
                        yshift=-7pt
                    },
                    point meta min=0,  
                    point meta max=100  
                ]
                \addplot[
                    matrix plot,
                    mesh/cols=3,
                    point meta=explicit,draw=gray
                ] table [meta=C] {
                    x y C
                    0 0 47
                    1 0 24
                    2 0 29

                    0 1 17
                    1 1 66
                    2 1 17

                    0 2 4
                    1 2 2
                    2 2 94

                };
            \end{axis}
        \end{tikzpicture}
    \caption*{Grad-based}
    \label{fig:confusion_naive+}
    \end{subfigure}
    \hfill
    \begin{subfigure}[]{0.375\textwidth}
    \centering
        \begin{tikzpicture}
            \begin{axis}[
                    width=0.72\linewidth,  
                    height=0.50\linewidth,  
                    colormap={custom}{
                        color(0cm)=(white);
                        color(1cm)=(yellow);
                        color(2cm)=(green);
                        color(3cm)=(cyan);
                    },
                    xlabel=Predicted,
                    xlabel style={yshift=0pt},
                    ylabel=Actual,
                    ylabel style={yshift=3pt},
                    xticklabels={ 1,  2,  3},
                    xtick={0,...,2},
                    xtick style={draw=none},
                    yticklabels={ 1,  2,  3},
                    ytick={0,...,2},
                    ytick style={draw=none},
                    enlargelimits=false,
                    xticklabel style={},
                    colorbar,
                    nodes near coords={\pgfmathprintnumber\pgfplotspointmeta},
                    nodes near coords style={
                        yshift=-7pt
                    },
                    point meta min=0,  
                    point meta max=100  
                ]
                \addplot[
                    matrix plot,
                    mesh/cols=3,
                    point meta=explicit,draw=gray
                ] table [meta=C] {
                    x y C
                    0 0 52
                    1 0 17
                    2 0 31

                    0 1 18
                    1 1 57
                    2 1 25

                    0 2 5
                    1 2 2
                    2 2 93

                };
            \end{axis}
        \end{tikzpicture}
    \caption*{Grad-free (ours)}
    \label{fig:confusion_alg2}
    \end{subfigure}
    
    \caption{Stable Diffusion 2.1. Accuracy: 186/300, 207/300, 202/300}
    \label{fig:confusions_a}
    \end{subfigure}
    
    \begin{subfigure}[]{\textwidth}
    \begin{subfigure}[]{0.3\textwidth}
    \centering
        \begin{tikzpicture}
            \begin{axis}[
                    width=0.9\linewidth,  
                    height=0.625\linewidth,  
                    colormap={custom}{
                        color(0cm)=(white);
                        color(1cm)=(yellow);
                        color(2cm)=(green);
                        color(3cm)=(cyan);
                    },
                    xlabel=Predicted,
                    xlabel style={yshift=0pt},
                    ylabel=Actual,
                    ylabel style={yshift=3pt},
                    xticklabels={ 1,  2,  3},
                    xtick={0,...,2},
                    xtick style={draw=none},
                    yticklabels={ 1,  2,  3},
                    ytick={0,...,2},
                    ytick style={draw=none},
                    enlargelimits=false,
                    xticklabel style={},
                    nodes near coords={\pgfmathprintnumber\pgfplotspointmeta},
                    nodes near coords style={
                        yshift=-7pt
                    },
                    point meta min=0,  
                    point meta max=100  
                ]
                \addplot[
                    matrix plot,
                    mesh/cols=3,
                    point meta=explicit,draw=gray
                ] table [meta=C] {
                    x y C
                    0 0 29
                    1 0 36
                    2 0 35

                    0 1 1
                    1 1 36
                    2 1 63

                    0 2 0
                    1 2 16
                    2 2 84

                };
            \end{axis}
        \end{tikzpicture}
    \caption*{Encoder}
    \label{fig:confusion_naive}
    \end{subfigure}
    \hfill
    \begin{subfigure}[]{0.3\textwidth}
        \centering
        \begin{tikzpicture}
            \begin{axis}[
                    width=0.9\linewidth,  
                    height=0.625\linewidth,  
                    colormap={custom}{
                        color(0cm)=(white);
                        color(1cm)=(yellow);
                        color(2cm)=(green);
                        color(3cm)=(cyan);
                    },
                    xlabel=Predicted,
                    xlabel style={yshift=0pt},
                    ylabel=Actual,
                    ylabel style={yshift=3pt},
                    xticklabels={ 1,  2,  3},
                    xtick={0,...,2},
                    xtick style={draw=none},
                    yticklabels={ 1,  2,  3},
                    ytick={0,...,2},
                    ytick style={draw=none},
                    enlargelimits=false,
                    xticklabel style={},
                    nodes near coords={\pgfmathprintnumber\pgfplotspointmeta},
                    nodes near coords style={
                        yshift=-7pt
                    },
                    point meta min=0,  
                    point meta max=100  
                ]
                \addplot[
                    matrix plot,
                    mesh/cols=3,
                    point meta=explicit,draw=gray
                ] table [meta=C] {
                    x y C
                    0 0 31
                    1 0 55
                    2 0 14

                    0 1 1
                    1 1 96
                    2 1 3

                    0 2 0
                    1 2 0
                    2 2 100

                };
            \end{axis}
        \end{tikzpicture}
    \caption*{Grad-based}
    \label{fig:confusion_naive+}
    \end{subfigure}
    \hfill
    \begin{subfigure}[]{0.375\textwidth}
    \centering
        \begin{tikzpicture}
            \begin{axis}[
                    width=0.72\linewidth,  
                    height=0.50\linewidth,  
                    colormap={custom}{
                        color(0cm)=(white);
                        color(1cm)=(yellow);
                        color(2cm)=(green);
                        color(3cm)=(cyan);
                    },
                    xlabel=Predicted,
                    xlabel style={yshift=0pt},
                    ylabel=Actual,
                    ylabel style={yshift=3pt},
                    xticklabels={ 1,  2,  3},
                    xtick={0,...,2},
                    xtick style={draw=none},
                    yticklabels={ 1,  2,  3},
                    ytick={0,...,2},
                    ytick style={draw=none},
                    enlargelimits=false,
                    xticklabel style={},
                    colorbar,
                    nodes near coords={\pgfmathprintnumber\pgfplotspointmeta},
                    nodes near coords style={
                        yshift=-7pt
                    },
                    point meta min=0,  
                    point meta max=100  
                ]
                \addplot[
                    matrix plot,
                    mesh/cols=3,
                    point meta=explicit,draw=gray
                ] table [meta=C] {
                    x y C
                    0 0 30
                    1 0 60
                    2 0 10

                    0 1 1
                    1 1 97
                    2 1 2

                    0 2 0
                    1 2 0
                    2 2 100

                };
            \end{axis}
        \end{tikzpicture}
    \caption*{Grad-free (ours)}
    \label{fig:confusion_alg2}
    \end{subfigure}
    
    \caption{InstaFlow. Accuracy: 149/300, 227/300, 227/300}
    \label{fig:confusions_b}
    \end{subfigure}
    \caption{Confusion matrices for watermark classification on LDMs. Ours is better than the encoder and works as well as the grad-based method~\citep{hong2023exact}.}
    \label{fig:confusions}    
\end{figure*}

\begin{figure*}[h]
    \small
    \centering
    \begin{minipage}{\textwidth}
    \setlength{\tabcolsep}{0pt}
    \begin{tabular}{@{}c@{ }c@{ }c@{ }c@{ }c@{ }c@{ }c@{ }c@{ }c@{ }c@{ }}
     \begin{tabular}{@{}c@{}}Embedded\\ watermark\end{tabular} & \begin{tabular}{@{}c@{}}Generated \end{tabular} & \multicolumn{2}{c@{ }}{\cellcolor{bGray} \begin{tabular}{@{}c@{ }} Encoder \\ (Recon. / Error)\end{tabular}} &  \multicolumn{2}{c@{ }}{\cellcolor{bGray} \begin{tabular}{@{}c@{}} Grad \\ (Recon. / Error) \end{tabular}} &  \multicolumn{2}{c@{ }}{\cellcolor{bGray} \begin{tabular}{@{}c@{}} Grad(short) \\ (Recon. / Error) \end{tabular}} & 
     \multicolumn{2}{c@{ }}{\cellcolor{bRed} \begin{tabular}{@{}c@{}} Grad-free 16bit (ours) \\ (Recon. / Error) \end{tabular}}  
     \\

     \multicolumn{2}{c@{}}{\# iter, Runtime} & \multicolumn{2}{c@{ }}{\cellcolor{bGray}-, -} & \multicolumn{2}{c@{ }}{\cellcolor{bGray}200, 64.4s} & \multicolumn{2}{c@{ }}{\cellcolor{bGray}100, 32.0s} & \multicolumn{2}{c@{ }}{\cellcolor{bRed}200, 35.0s} \\

     \includegraphics[width=0.092\textwidth]{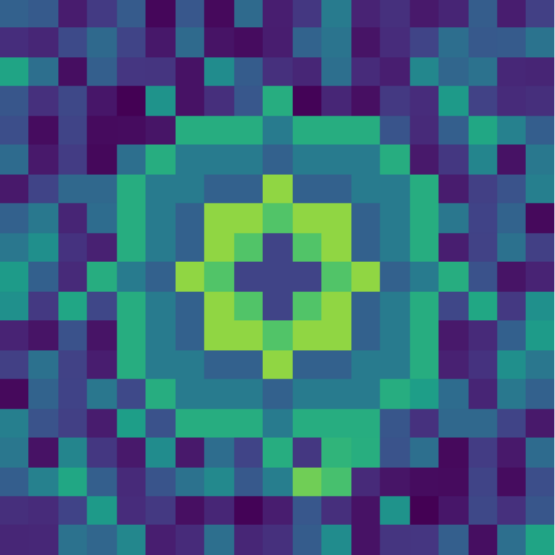} & \includegraphics[width=0.092\textwidth]{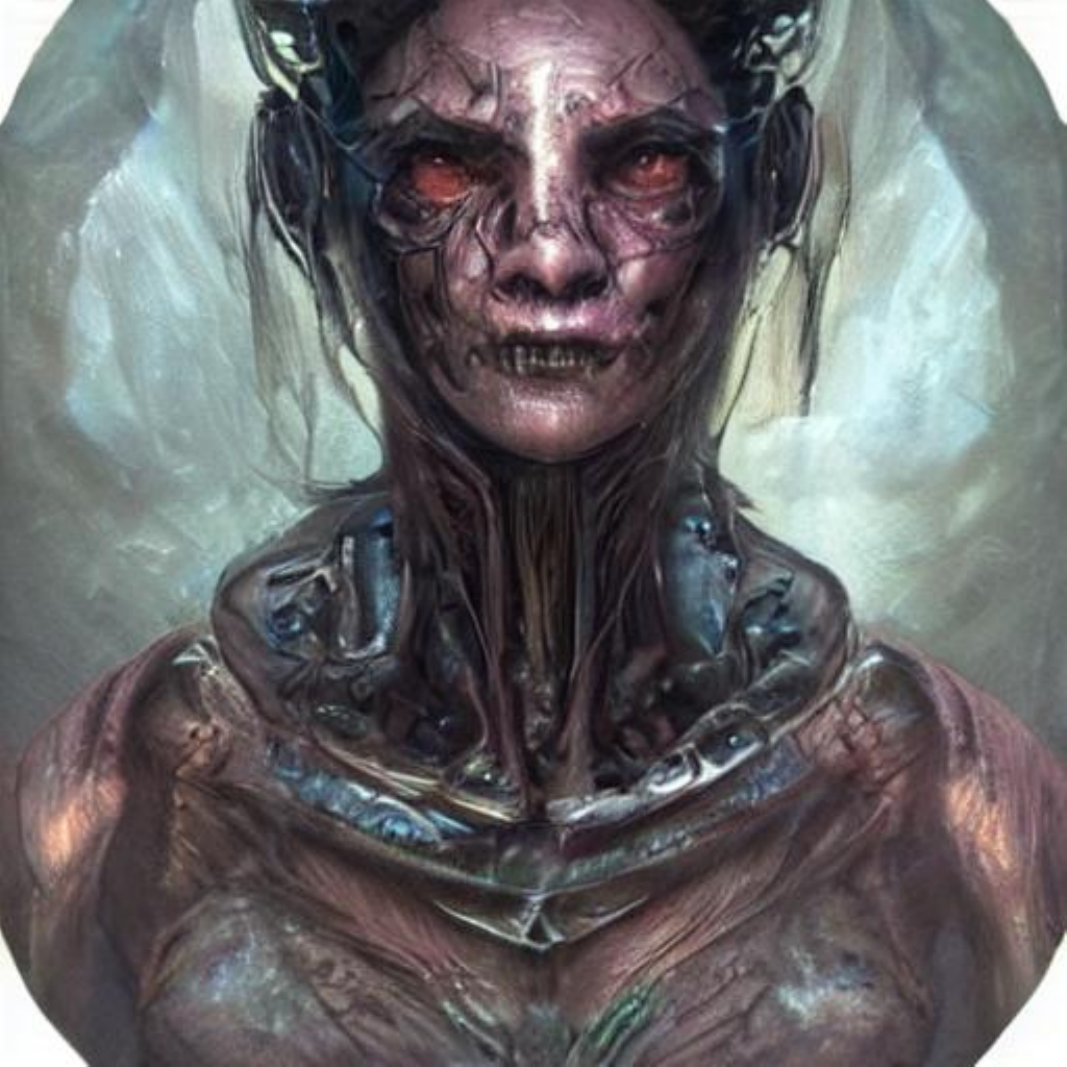} & \includegraphics[width=0.092\textwidth]{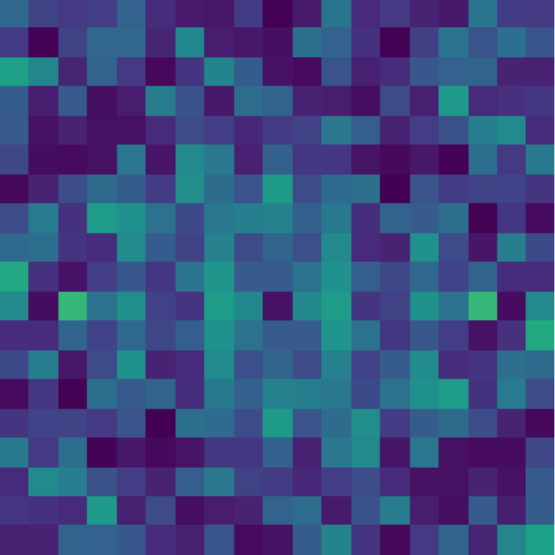} & \includegraphics[width=0.092\textwidth]{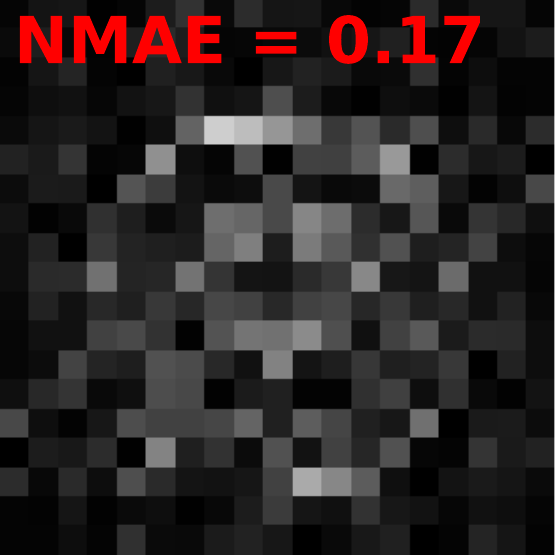} & \includegraphics[width=0.092\textwidth]{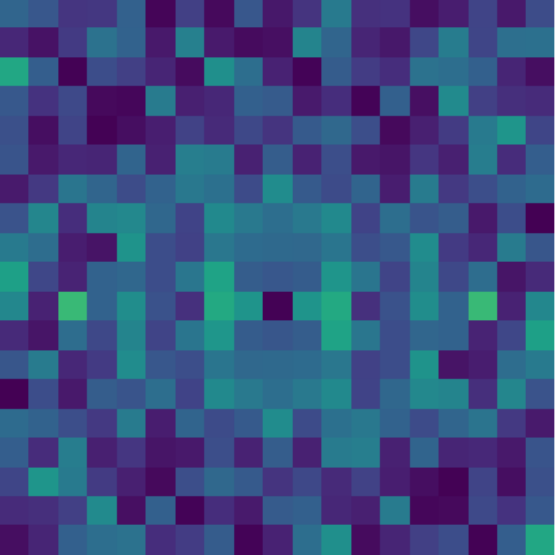} & \includegraphics[width=0.092\textwidth]{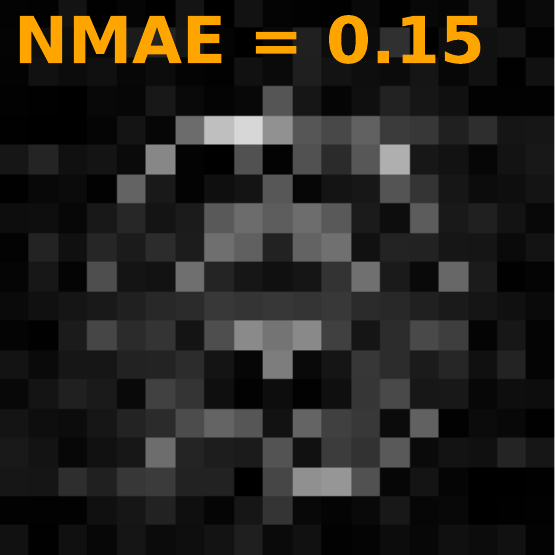} & \includegraphics[width=0.092\textwidth]{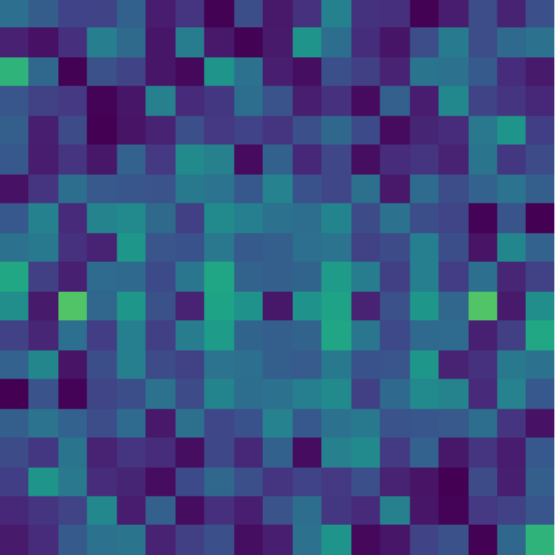} & \includegraphics[width=0.092\textwidth]{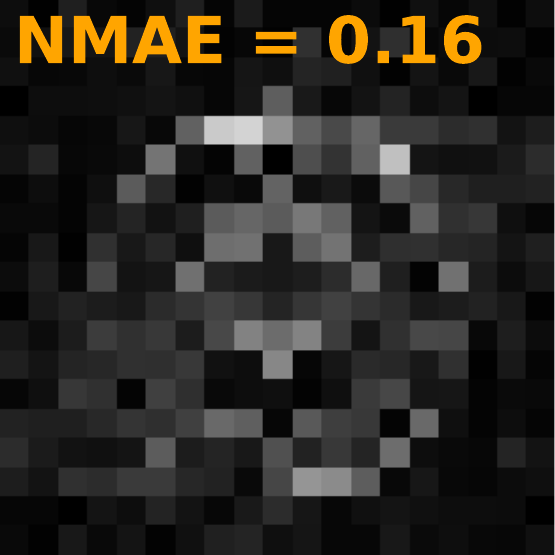} & \includegraphics[width=0.092\textwidth]{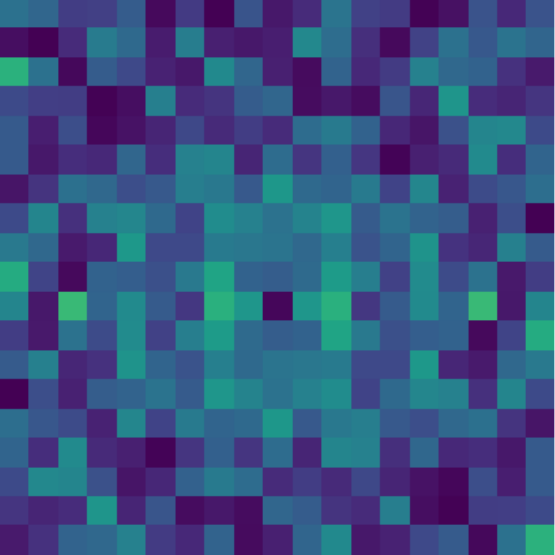} & \includegraphics[width=0.092\textwidth]{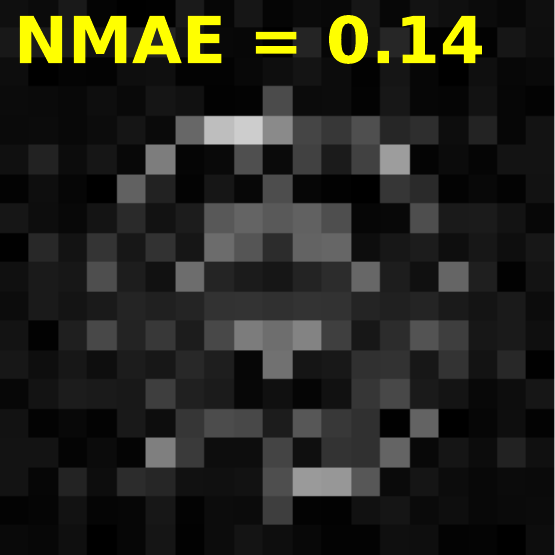} \\

     \includegraphics[width=0.092\textwidth]{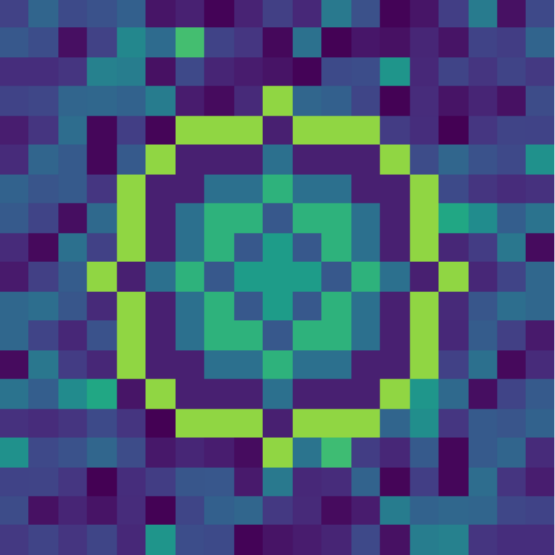} & \includegraphics[width=0.092\textwidth]{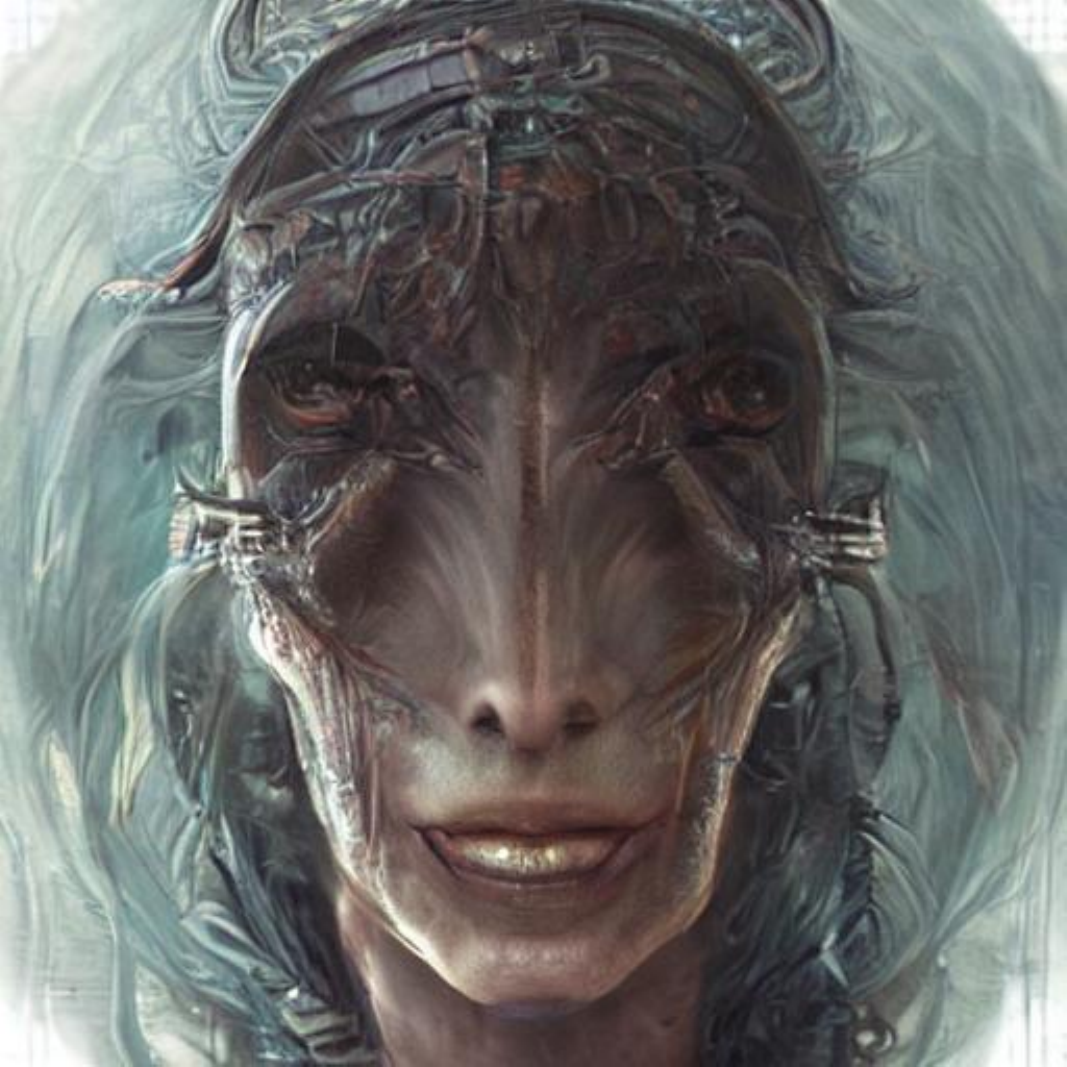} & \includegraphics[width=0.092\textwidth]{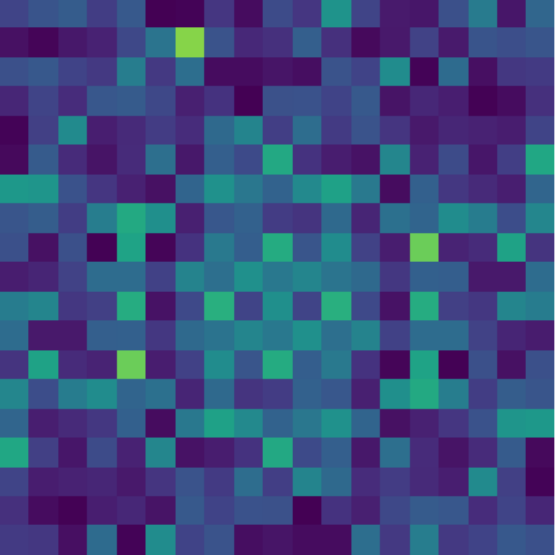} & \includegraphics[width=0.092\textwidth]{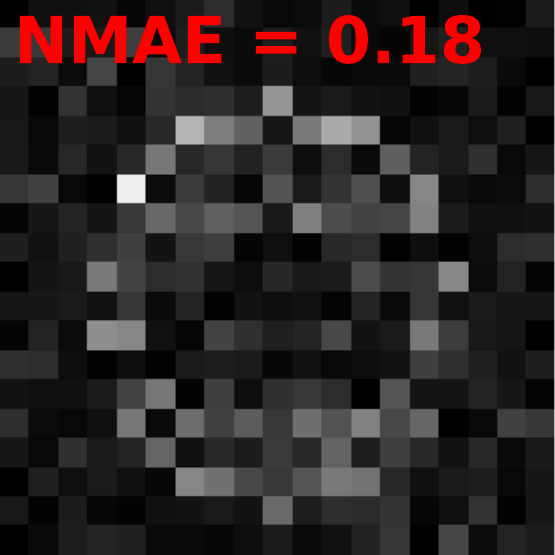} & \includegraphics[width=0.092\textwidth]{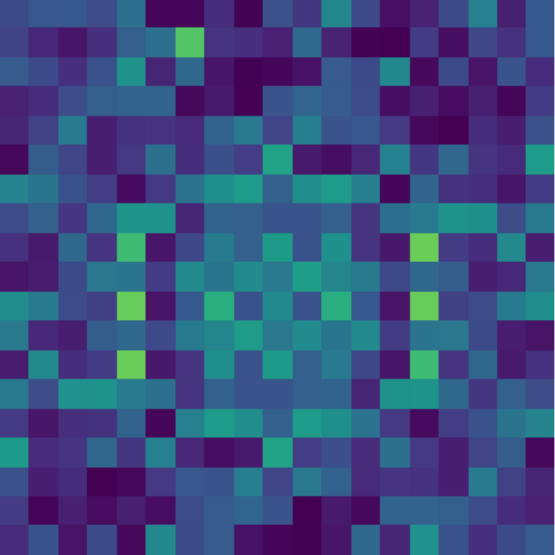} & \includegraphics[width=0.092\textwidth]{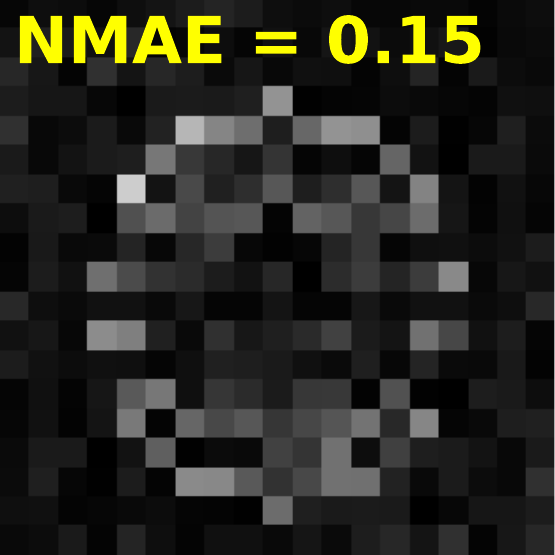} & \includegraphics[width=0.092\textwidth]{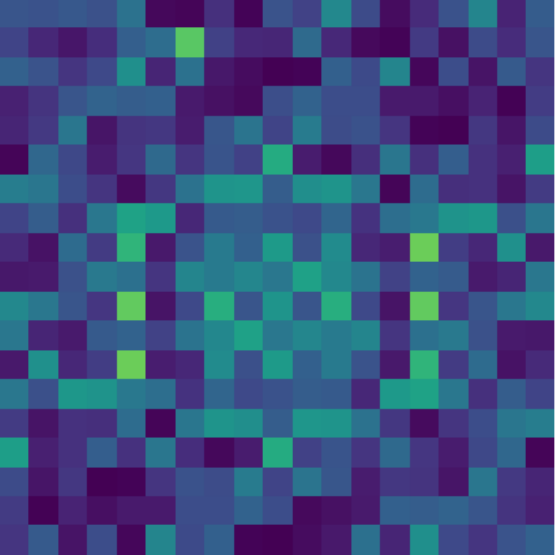} & \includegraphics[width=0.092\textwidth]{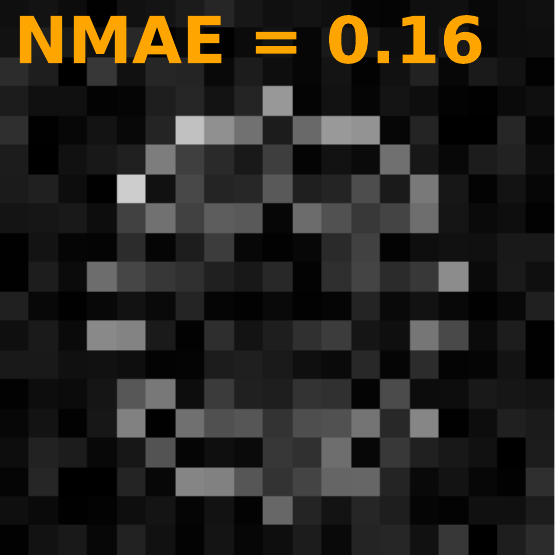} & \includegraphics[width=0.092\textwidth]{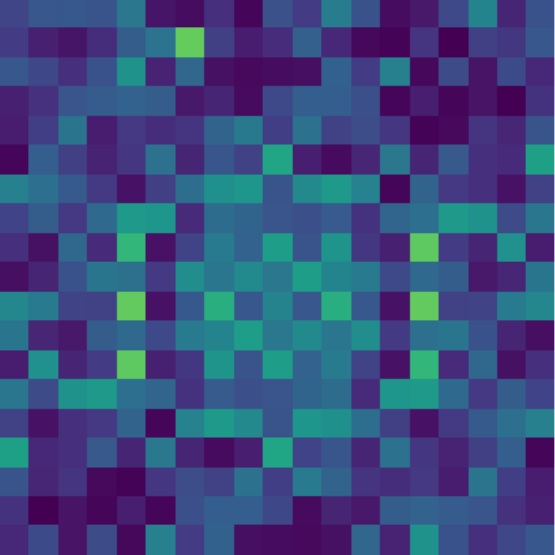} & \includegraphics[width=0.092\textwidth]{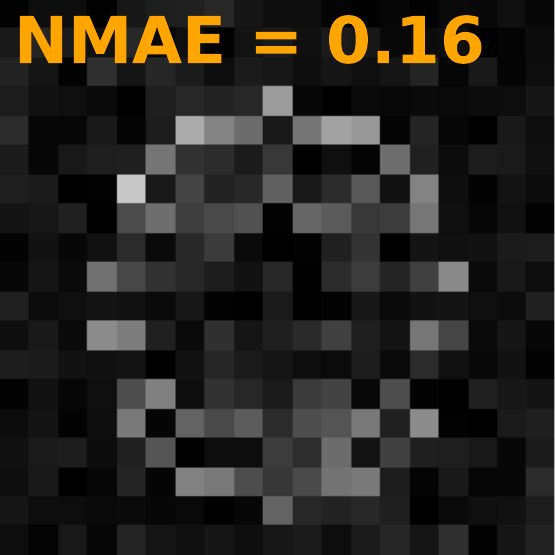} \\
    
    \includegraphics[width=0.092\textwidth]{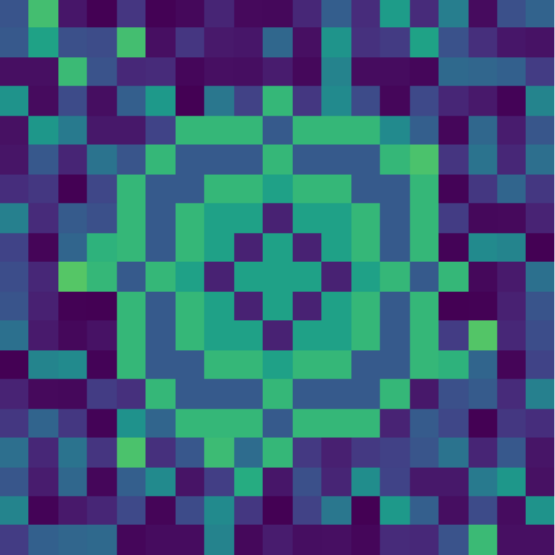} & \includegraphics[width=0.092\textwidth]{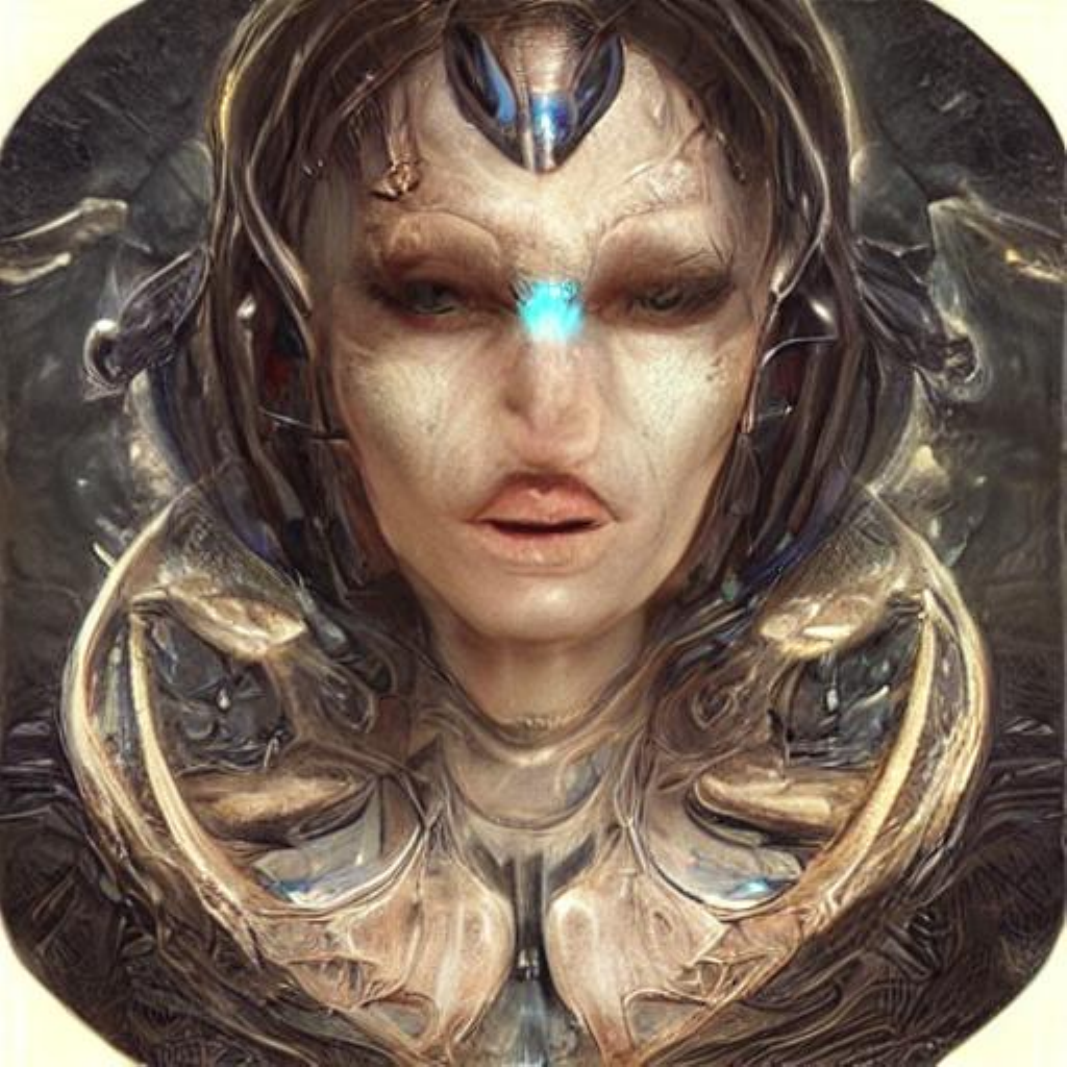} & \includegraphics[width=0.092\textwidth]{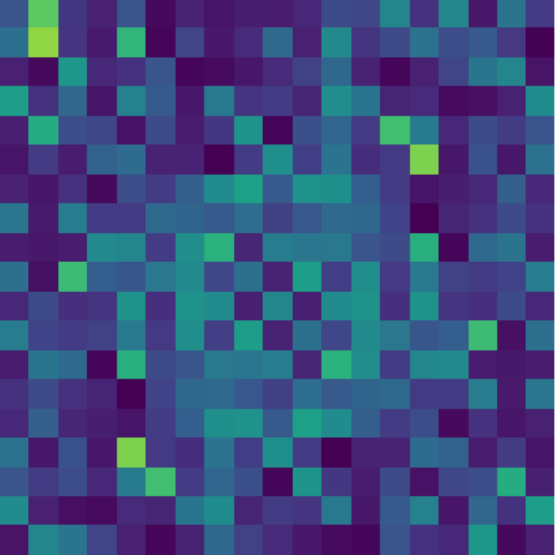} & \includegraphics[width=0.092\textwidth]{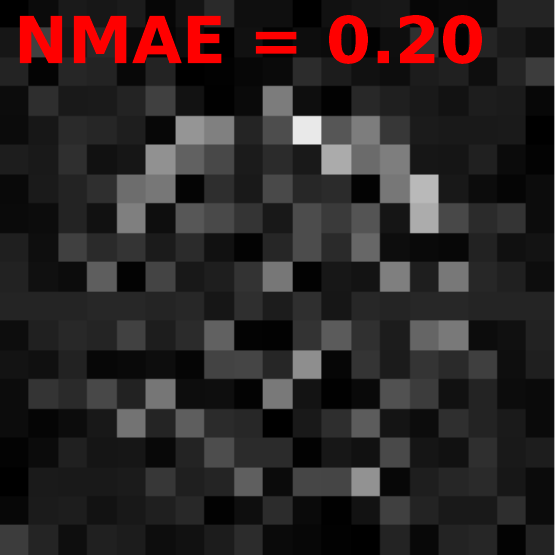} & \includegraphics[width=0.092\textwidth]{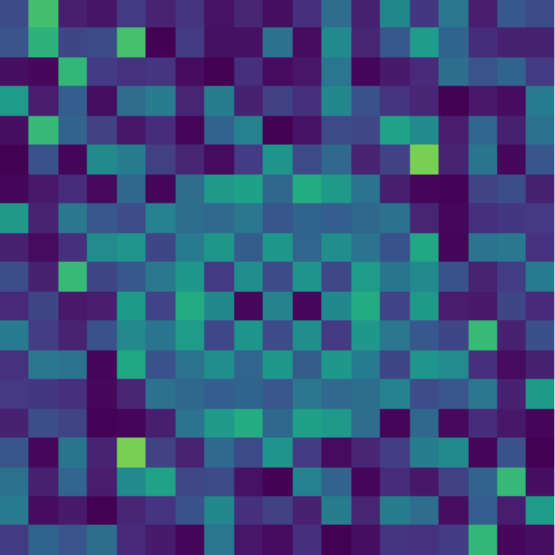} & \includegraphics[width=0.092\textwidth]{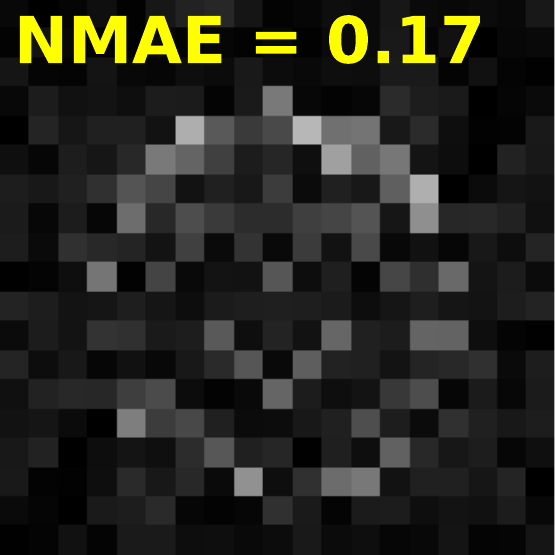} & \includegraphics[width=0.092\textwidth]{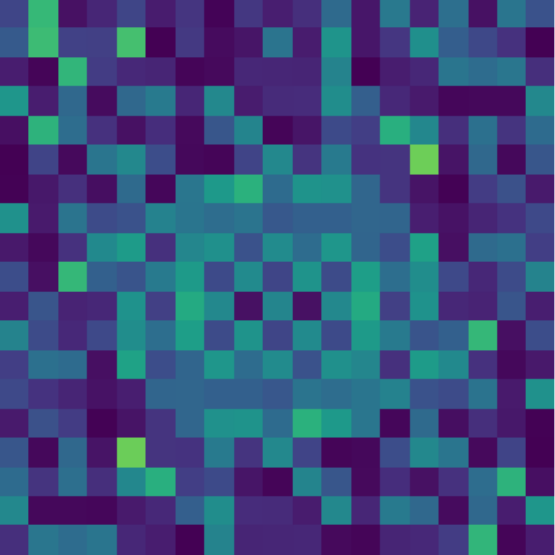} & \includegraphics[width=0.092\textwidth]{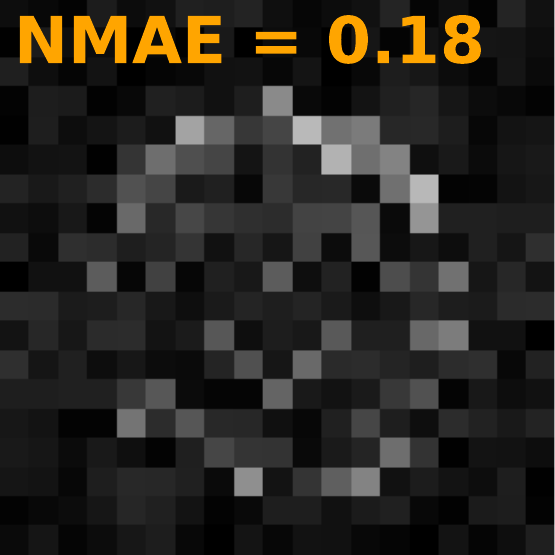} & \includegraphics[width=0.092\textwidth]{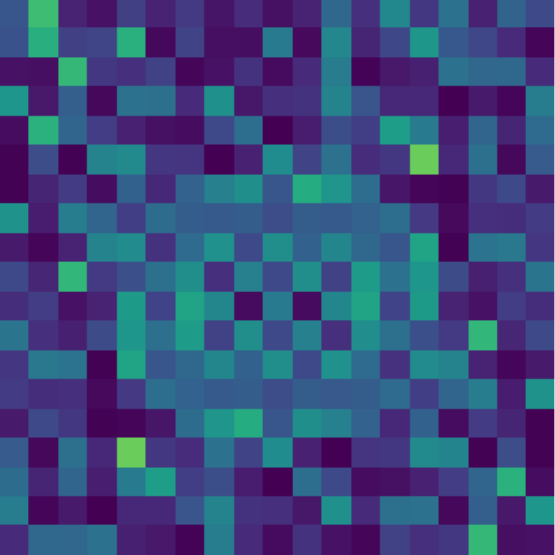} & \includegraphics[width=0.092\textwidth]{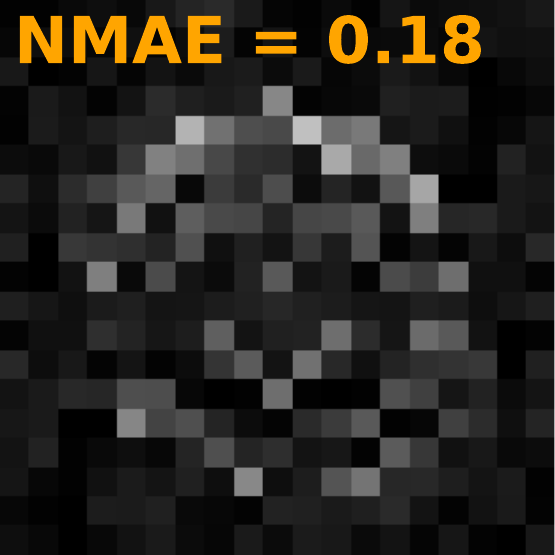} \\
     
    \end{tabular}
    \end{minipage}
    \caption{\textcolor{black}{Qualitative result on the watermarking classification experiment ~\citep{hong2023exact}. Our grad-free decoder inversion aids in the accurate reconstruction of Tree-ring watermarking~\cite{wen2023tree}. Specifically, it either reduces the processing time compared to the grad-based (column 3 vs 5) or achieves better accuracy within the same runtime (column 4 vs 5).}}
    \label{fig:watermarks2}
\end{figure*}

\begin{figure*}[h]
    \setlength{\tabcolsep}{1pt}
    \centering
    \begin{tabular}{@{}l@{}c@{  }c@{}c@{  }c@{}c@{}}
    Method & \begin{tabular}{@{}c@{}} \scriptsize \# iter \\ \scriptsize Runtime \end{tabular} & Edited & Error map ($\times 5$) & Edited & Error map ($\times 5$) \\
     \begin{tabular}{@{}l@{}} \small Oracle \\ {[32]} \end{tabular} & - &
    \begin{tabular}{c}\includegraphics[ clip, width=0.211\linewidth]{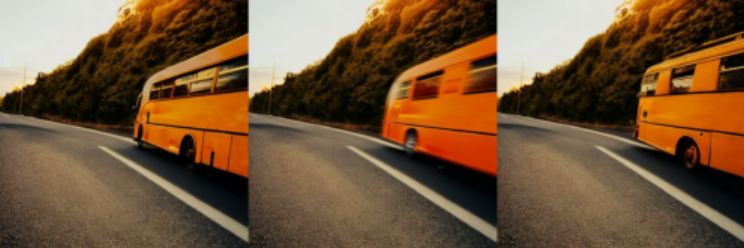}\end{tabular} & \begin{tabular}{c}\includegraphics[clip, width=0.211\linewidth]{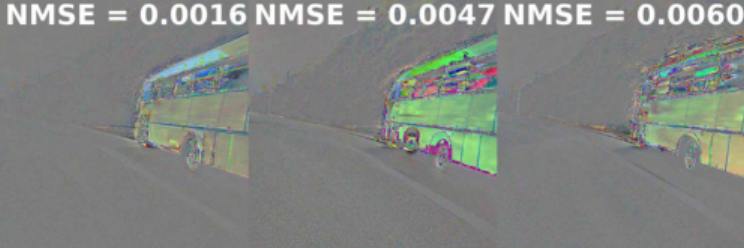}\end{tabular}& \begin{tabular}{c}\includegraphics[clip, width=0.211\linewidth]{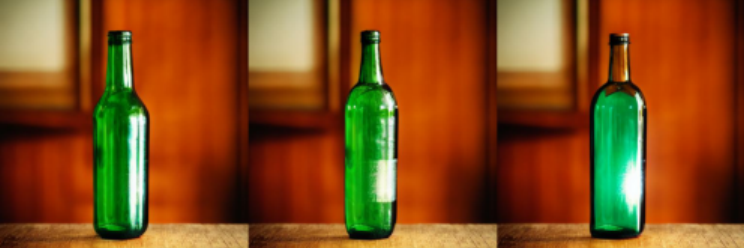}\end{tabular} & \begin{tabular}{c}\includegraphics[clip, width=0.211\linewidth]{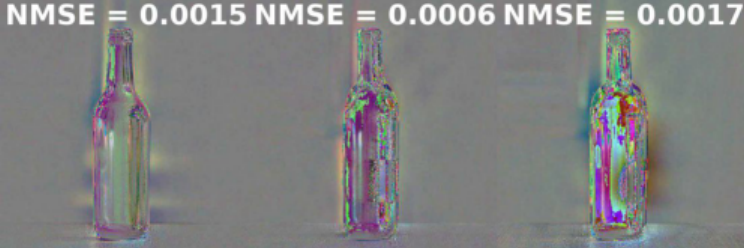}\end{tabular}\\
    \cellcolor{bGray} \begin{tabular}{@{}l@{}} \small Grad-\\based \end{tabular} &  \cellcolor{bGray} \begin{tabular}{@{}c@{}} 50 \\ 15.7s \end{tabular} & \begin{tabular}{c}\includegraphics[clip, width=0.211\linewidth]{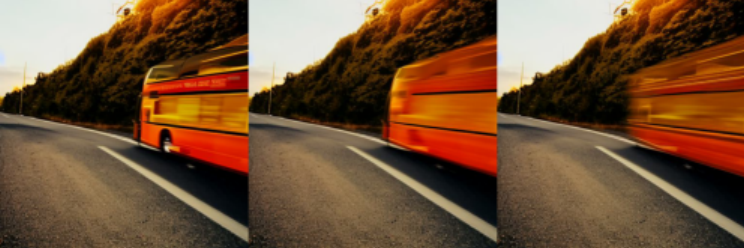}\\[-\dp\strutbox]\end{tabular}&\begin{tabular}{c}\includegraphics[clip, width=0.211\linewidth]{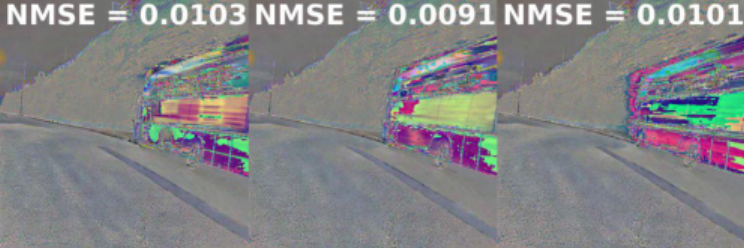}\\[-\dp\strutbox]\end{tabular}& \begin{tabular}{c}\includegraphics[clip, width=0.211\linewidth]{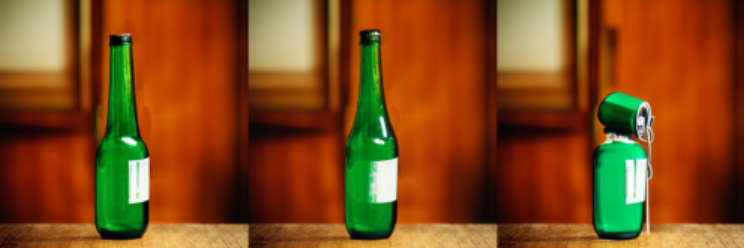}\\[-\dp\strutbox]\end{tabular} & \begin{tabular}{c}\includegraphics[clip, width=0.211\linewidth]{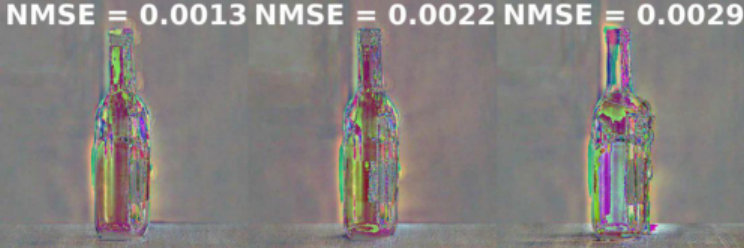}\\[-\dp\strutbox]\end{tabular} \\ 
    \cellcolor{bBlue} \begin{tabular}{@{}l@{}} \scriptsize Grad-free \\ \scriptsize 32bit \\ \scriptsize (Ours) \end{tabular}  &  \cellcolor{bBlue} \begin{tabular}{@{}c@{}} 100 \\ 17.7s \end{tabular} & \begin{tabular}{c}\includegraphics[ clip, width=0.211\linewidth]{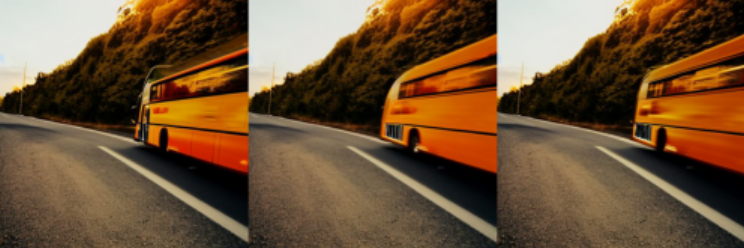}\\[-\dp\strutbox]\end{tabular}&\begin{tabular}{c} \includegraphics[clip, width=0.211\linewidth]{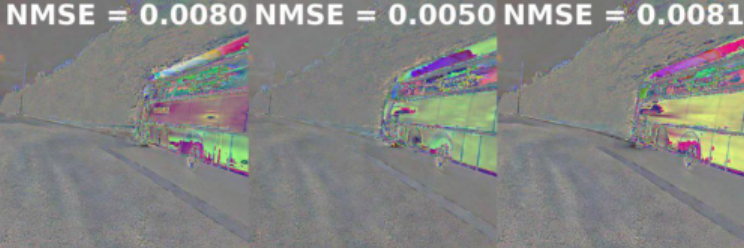}\\[-\dp\strutbox]\end{tabular}& \begin{tabular}{c}\includegraphics[clip, width=0.211\linewidth]{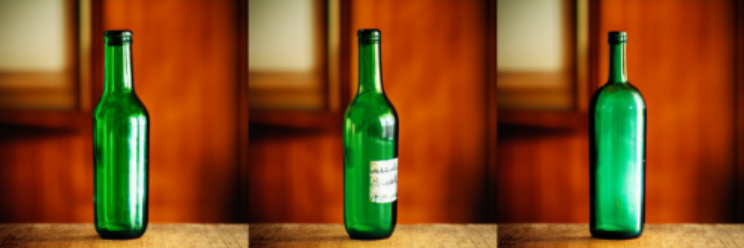}\\[-\dp\strutbox]\end{tabular} & \begin{tabular}{c}\includegraphics[clip, width=0.211\linewidth]{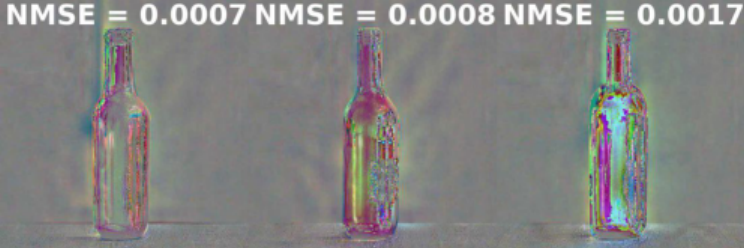}\\[-\dp\strutbox]\end{tabular}\\
    \cellcolor{bBlue} \begin{tabular}{@{}l@{}} \scriptsize Grad-free \\ \scriptsize 16bit \\ \scriptsize (Ours) \end{tabular}  &  \cellcolor{bBlue} 
 \begin{tabular}{@{}c@{}} 200 \\ 19.3s \end{tabular} & \begin{tabular}{c}\includegraphics[ clip, width=0.211\linewidth]{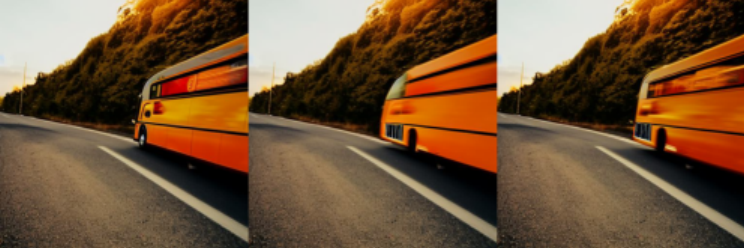}\\[-\dp\strutbox]\end{tabular}&\begin{tabular}{c} \includegraphics[clip, width=0.211\linewidth]{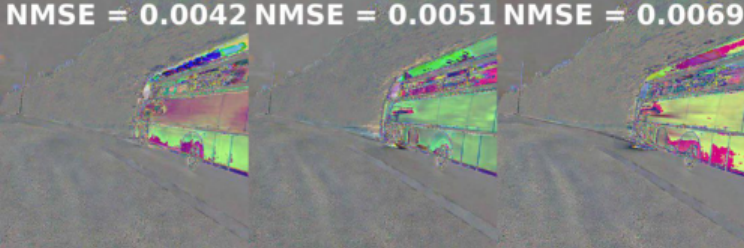}\\[-\dp\strutbox]\end{tabular}& \begin{tabular}{c}\includegraphics[clip, width=0.211\linewidth]{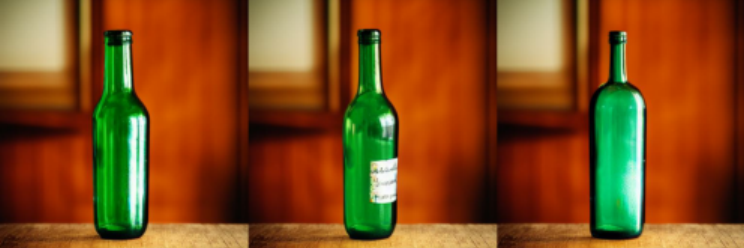}\\[-\dp\strutbox]\end{tabular} & \begin{tabular}{c}\includegraphics[clip, width=0.211\linewidth]{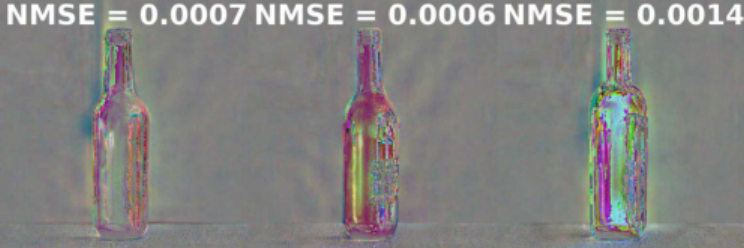}\\[-\dp\strutbox]\end{tabular}\\    \vspace{-10pt} \\
    \cellcolor{bGray} \begin{tabular}{@{}l@{}} \small Grad-\\based \end{tabular} &  \cellcolor{bGray} \begin{tabular}{@{}c@{}} 30 \\ 9.31s \end{tabular} & \begin{tabular}{c}\includegraphics[clip, width=0.211\linewidth]{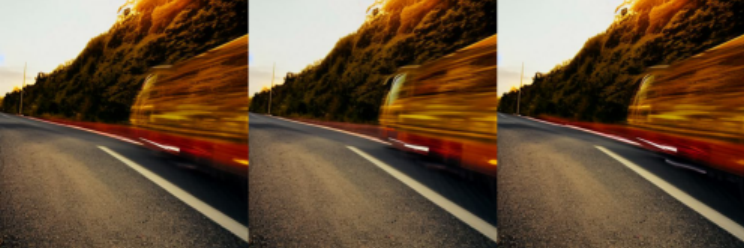}\\[-\dp\strutbox]\end{tabular}&\begin{tabular}{c}\includegraphics[clip, width=0.211\linewidth]{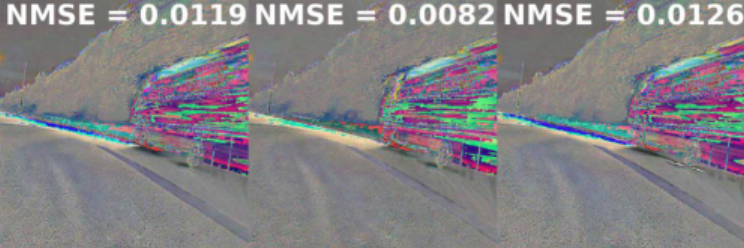}\\[-\dp\strutbox]\end{tabular}& \begin{tabular}{c}\includegraphics[clip, width=0.211\linewidth]{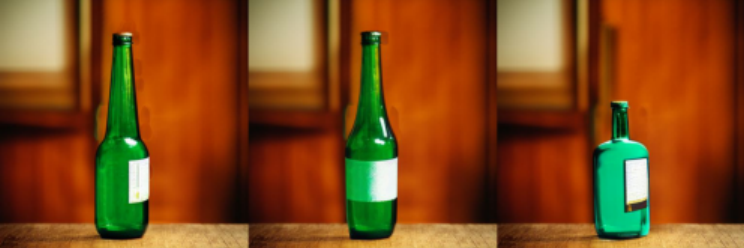}\\[-\dp\strutbox]\end{tabular} & \begin{tabular}{c}\includegraphics[clip, width=0.211\linewidth]{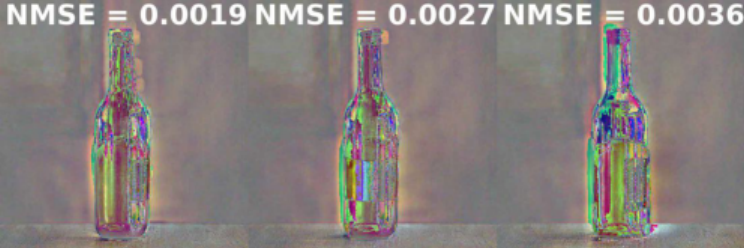}\\[-\dp\strutbox]\end{tabular} \\ 
    \cellcolor{bBlue} \begin{tabular}{@{}l@{}} \scriptsize Grad-free \\ \scriptsize 32bit \\ \scriptsize (Ours) \end{tabular}  & \cellcolor{bBlue} \begin{tabular}{@{}c@{}} 50 \\ 8.53s \end{tabular} & \begin{tabular}{c}\includegraphics[ clip, width=0.211\linewidth]{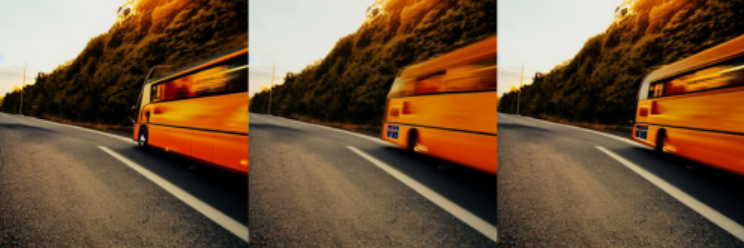}\\[-\dp\strutbox]\end{tabular}&\begin{tabular}{c} \includegraphics[clip, width=0.211\linewidth]{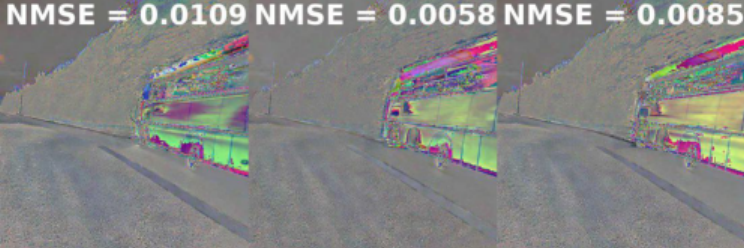}\\[-\dp\strutbox]\end{tabular}& \begin{tabular}{c}\includegraphics[clip, width=0.211\linewidth]{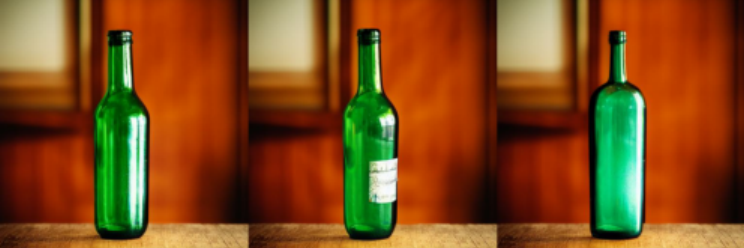}\\[-\dp\strutbox]\end{tabular} & \begin{tabular}{c}\includegraphics[clip, width=0.211\linewidth]{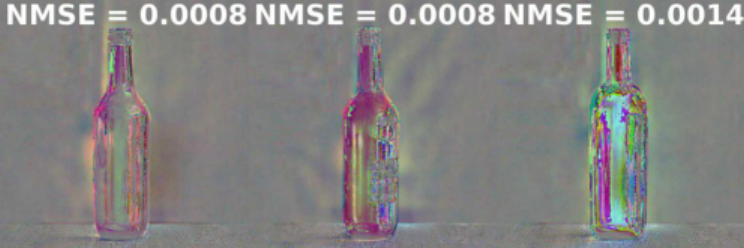}\\[-\dp\strutbox]\end{tabular}\\
    \cellcolor{bBlue} \begin{tabular}{@{}l@{}} \scriptsize Grad-free \\ \scriptsize 16bit \\ \scriptsize (Ours) \end{tabular}  & \cellcolor{bBlue}\begin{tabular}{@{}c@{}} 100 \\ 9.56s \end{tabular} & \begin{tabular}{c}\includegraphics[ clip, width=0.211\linewidth]{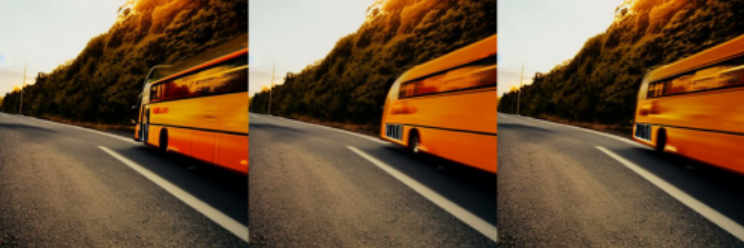}\\[-\dp\strutbox]\end{tabular}&\begin{tabular}{c} \includegraphics[clip, width=0.211\linewidth]{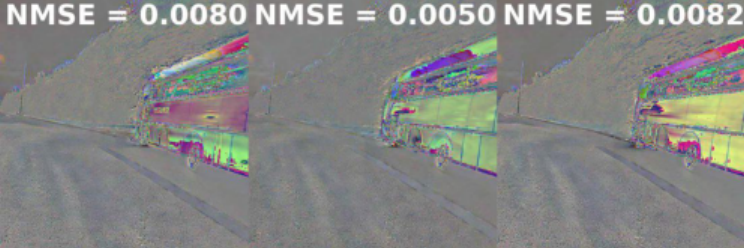}\\[-\dp\strutbox]\end{tabular}& \begin{tabular}{c}\includegraphics[clip, width=0.211\linewidth]{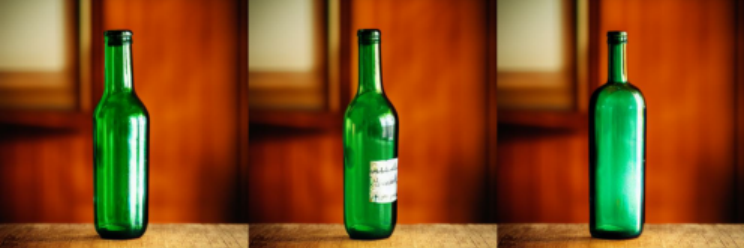}\\[-\dp\strutbox]\end{tabular} & \begin{tabular}{c}\includegraphics[clip, width=0.211\linewidth]{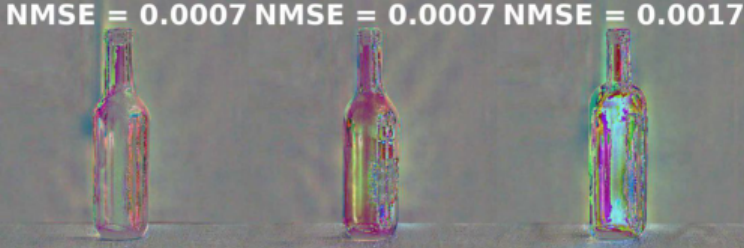}\\[-\dp\strutbox]\end{tabular}  \\
    \end{tabular}
    \caption{
    \textcolor{black}{Our grad-free methods enables the background-preserving image editing, where the original trajectory $(\vz_t)$ is unknown. The first row (Oracle) shows the result with using the original trajectory. The latter cases estimate the trajectory with inversion methods. Our grad-free method demonstrates better performance compared to the grad-based method at a similar runtime. 
    }}
    \label{fig:editing}
\end{figure*}

\textcolor{black}{\subsubsection{Another application: Background-preserving image editing}
Background-preserving editing, to manipulate an image based on a new condition while preserving background from the original image, requires the entire generating latent trajectory. When it is unknown, it must be recovered via inversion~\citep{patashnik2023localizing}. We empirically demonstrate that our algorithm enhances the background-preserving image editing, without the need for the original latents. \Cref{fig:editing} shows the qualitative results of applying our algorithm to the experiment. To compare accuracy at similar execution times, we adjusted the number of iterations to match the execution time. At comparable execution times, our grad-free method better preserves the background and achieves a lower NMSE.
}

\textcolor{black}{\subsubsection{Ablation Studies}
We conducted an ablation study on each optimizer and learning rate scheduling, in SD2.1(32bit). ~\Cref{tab:ablationa} shows the result of only changing the optimizer while keeping all other conditions the same. In ~\Cref{tab:ablationb}, we observed what happens when using a fixed learning rate instead of applying learning rate scheduling with Adam. When using a fixed learning rate, we found that with a large learning rate (lr=0.01), the performance was poor when the number of iterations was high, and with a small learning rate (lr=0.002), the performance was poor when the number of iterations was low. In contrast, the scheduled learning rate we used showed consistent performance across all intervals regardless of the number of iterations.
}

\begin{table}[!t]
\caption{\textcolor{black}{Ablation studies on optimizer and learning rate scheduling. Adam showed the best result compared to vanilla and KM iterations method. The learning scheduling we used showed consistently good performance across all intervals.}}
\label{tab:ablation}
\centering
\begin{subtable}{\linewidth}
\centering
\caption{Ablation study on optimizer}\label{tab:ablationa}
\begin{tabular}{@{}c|cccc@{}}
\toprule
 & \multicolumn{4}{c}{NMSE (dB)}  \\ 
\cmidrule{2-5}
   Method $\backslash$ \# iter. & 20 & 30 & 50 & 100 \\  \midrule
Vanilla & -16.87 $\pm$ 0.38 & -17.42 $\pm$ 0.41 & -18.21 $\pm$ 0.46 & -19.35 $\pm$ 0.54 \\ KM iterations & -18.99 $\pm$ 0.53 & -20.72 $\pm$ 0.73 & -21.46 $\pm$ 0.96 & -20.91 $\pm$ 1.20 \\
Adam (orig.) & \textbf{-19.39} $\pm$ 0.54 & \textbf{-20.84} $\pm$ 0.66 & \textbf{-21.71} $\pm$ 0.77 & \textbf{-21.85} $\pm$ 0.82 \\ 
\bottomrule
\end{tabular}
\end{subtable}

\hfill

\begin{subtable}{\linewidth}
\centering
\caption{Ablation study on learning rate scheduling}\label{tab:ablationb}
\begin{tabular}{@{}c|cccc@{}}
\toprule
\multirow{2}{*}{Method} & \multicolumn{4}{c}{NMSE (dB)}  \\ 
\cmidrule{2-5}
& 20 & 50 & 100 & 200 \\  \midrule
lr=0.01 (fixed) & \textbf{-20.05} $\pm$ 0.58 & \textbf{-21.07} $\pm$ 0.70 & -21.22 $\pm$ 0.74 & -20.61 $\pm$ 0.79 \\ 
        lr=0.002 (fixed) & -17.85 $\pm$ 0.43 & -19.28 $\pm$ 0.53 & -20.59 $\pm$ 0.64 & -21.57 $\pm$ 0.74 \\ 
        lr scheduled (orig.) & -19.39 $\pm$ 0.54 & -20.84 $\pm$ 0.66 & \textbf{-21.71} $\pm$ 0.77 & \textbf{-21.85} $\pm$ 0.82 \\
\bottomrule
\end{tabular}
\end{subtable}

\end{table}

\begin{figure*}
    \centering
          \begin{subfigure}{0.32\textwidth}
            \centering
            \includegraphics[width=\textwidth]{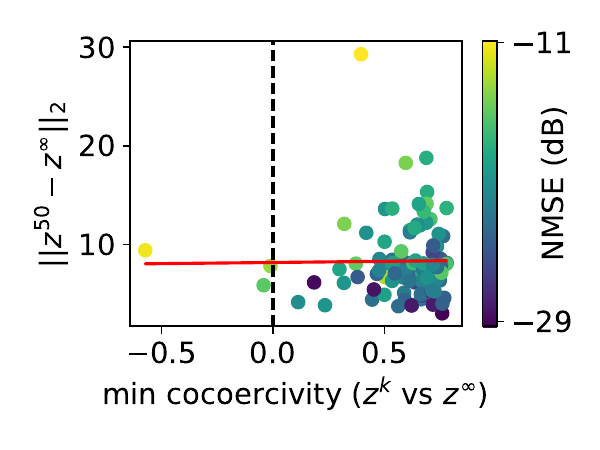} 
            \caption{Stable Diffusion 2.1}
            \label{fig:cocoercivity_adam_1}
          \end{subfigure}
          \begin{subfigure}{0.32\textwidth}
            \centering
            \includegraphics[width=\textwidth]{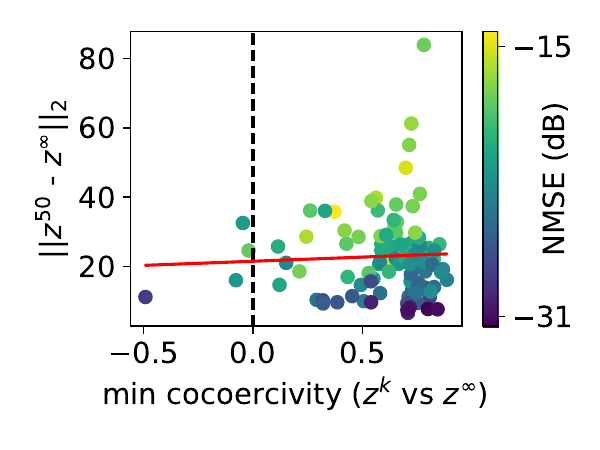} 
            \caption{LaVie}
            \label{fig:cocoercivity_adam_2}
          \end{subfigure}
          \begin{subfigure}{0.32\textwidth}
            \centering
            \includegraphics[width=\textwidth]{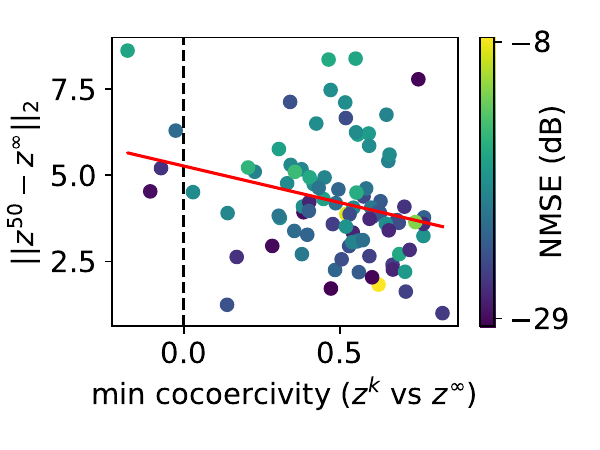} 
            \caption{InstaFlow}
            \label{fig:cocoercivity_adam_3}
          \end{subfigure}
          \caption{\textcolor{black}{Analysis on the grad-based method. x-axis: $\underset{k\in[0,50]}{\min}\frac{\langle \gE \gD \vz^{\infty} - \gE \gD \vz^{k}, \vz^{\infty} - \vz^{k} \rangle }{ \lVert \gE \gD \vz^{\infty} - \gE \gD \vz^{k} \rVert_2^2}$, y-axis: convergence (\ie, $\lVert \vz^{50} - \vz^\infty \rVert_2$). The red line shows the linear function fitted by least squares. Note that $\vz^{\infty}$ was approximated by $\vz^{300}$. Better convergence and accuracy were observed, but no relationship with cocoercivity was found.}}
          \label{fig:cocoercivity_grad}
\end{figure*}

\textcolor{black}{\subsubsection{Analysis on the grad-based method}
In ~\Cref{fig:cocoercivity_grad}, we present the instance-wise cocoercivity, convergence, and accuracy for the gradient-based method, similar to ~\Cref{fig:cocoercivity} and ~\Cref{fig:cocoercivity_adam}. Similar to the results observed with the gradient-free method, the gradient-based method showed that most instances satisfied cocoercivity, and better convergence often led to higher accuracy. However, it was observed that cocoercivity and convergence are not significantly correlated. In other words, the correlation between cocoercivity and convergence is not a general characteristic, but a unique feature we discovered in our gradient-free method.
}

\subsection{Supplementary explanation}
\paragraph{\Cref{fig:generative_models}} Normalizing flows (NFs) are born with its analytic invertibility, hence located at the bottom left. VAEs and learning-based GAN inversion methods have corresponding encoders, so they have short computation time but come with some inversion error. Some special structured DMs~\citep{wallace2023edict} are also located here. 

\subsection{Computational resources}
For running Stable Diffusion 2.1, one NVIDIA GeForce RTX 3090 Ti was used. The RAM size of the GPU was 24 GB. Note that most of the computation was conducted on GPU. For CPU, one 11th Gen Intel(R) Core(TM) i9-11900KF @ 3.50GHz was used. For running LaVie, one NVIDIA A100 SXM4 80GB and AMD EPYC 7742 64-Core Processor were used. For InstaFlow, one NVIDIA GeForce RTX 3090 GPU with one 12th Gen Intel(R) Core(TM) 17-12700K @ 3.60GHz was used.


\end{document}